\documentclass[11pt]{article}

\usepackage[preprint]{acl}

\usepackage{times}
\usepackage{latexsym}
\usepackage{amsmath}
\usepackage{booktabs}
\usepackage{tabularx}
\usepackage{amssymb}
\usepackage{enumitem}
\usepackage{array}
\usepackage{tabularx}
\usepackage{algorithm}
\usepackage{algpseudocode}
\usepackage{listings}
\usepackage[most]{tcolorbox}
\usepackage{xcolor}

\newcolumntype{C}[1]{>{\centering\arraybackslash}m{#1}}
\newcolumntype{Y}{>{\centering\arraybackslash}X}

\newtcblisting{promptbox}[1]{
  listing only,
  enhanced,
  colback=white,
  colframe=black!65,
  boxrule=0.4pt,
  arc=1pt,
  left=3pt,
  right=3pt,
  top=3pt,
  bottom=3pt,
  before skip=4pt,
  after skip=4pt,
  title=\textbf{#1},
  fonttitle=\bfseries,
  listing options={
    basicstyle=\normalfont\normalsize,
    breaklines=true,
    breakatwhitespace=true,
    breakautoindent=false,
    breakindent=0pt,
    columns=fullflexible,
    keepspaces=true,
    showstringspaces=false,
    escapeinside={(*@}{@*)}
  }
}

\usepackage[T1]{fontenc}
\usepackage[utf8]{inputenc}

\usepackage{microtype}

\usepackage{inconsolata}

\usepackage{graphicx}
\usepackage{xurl}

\title{Beyond Objective Equivalence: Constraint Injection for LLM-Based Optimization Modeling on Vehicle Routing Problems}

\author{
  \textbf{Xizi Luo\textsuperscript{1,3}},
  \textbf{Changhong He\textsuperscript{2,3}},
  \textbf{Dongdong Geng\textsuperscript{3}},
  \textbf{Chenggong Shi\textsuperscript{3}},
  \textbf{Yu Mei\textsuperscript{3}}\thanks{Corresponding author: \texttt{whqyqy@hotmail.com}}
  \\
  \textsuperscript{1}School of Economics and Management, Beihang University, Beijing, China
  \\
  \textsuperscript{2}School of Mechanical Engineering and Automation, Beihang University, Beijing, China
  \\
  \textsuperscript{3}Baidu Inc., Beijing, China
}

\begin{document}
\maketitle
\begin{abstract}
Large language models (LLMs) can generate executable solver code from natural-language descriptions of optimization problems. However, existing verification signals focus on whether the generated formulation produces the correct objective value. Such signals can overlook constraint-level errors: incorrect formulations may still produce the same optimum when extra or missing constraints do not affect the tested instance. We propose constraint injection, a verification method that directly tests whether the generated code implements the intended constraints. We use feasible solutions to detect incorrectly added constraints and one-constraint-violating solutions to detect omissions. Together with the optimal value check, these tests verify both the objective and the constraint set. We evaluate this approach on vehicle routing problems (VRPs), which provide a challenging testbed with diverse, tightly coupled operational constraints. We develop VRPCoder, an 8B model for generating Gurobi code from natural-language VRP descriptions, and VRPBench, an expert-verified benchmark spanning 21 VRP variants across four subsets. The verifier serves as a rejection-sampling criterion for synthetic data construction and as a rollout-level reward for reinforcement learning (RL). On VRPBench, VRPCoder-RL reaches 93\% overall Pass@1, outperforms Gemini-3.1-Pro Preview on three subsets, exceeds Claude-Sonnet-4.5 by 28 percentage points, and exceeds the strongest prior OR-LLM by 78 percentage points.

\end{abstract}

\section{Introduction}
Optimization problems are often described in natural language, but solving them requires translating these descriptions into formal mathematical models and executable solver code~\citep{ramamonjison2023nl4opt}. Large language models (LLMs) provide a promising way to automate this process. However, verifying the correctness of the generated formulation remains difficult: code that runs successfully and produces the expected objective value may still fail to faithfully encode the intended constraints. Such errors are particularly concerning in optimization modeling, where an incorrect constraint set can yield seemingly valid solutions while misrepresenting the original problem.

\begin{figure}[t]
    \centering
    \includegraphics[width=\columnwidth]{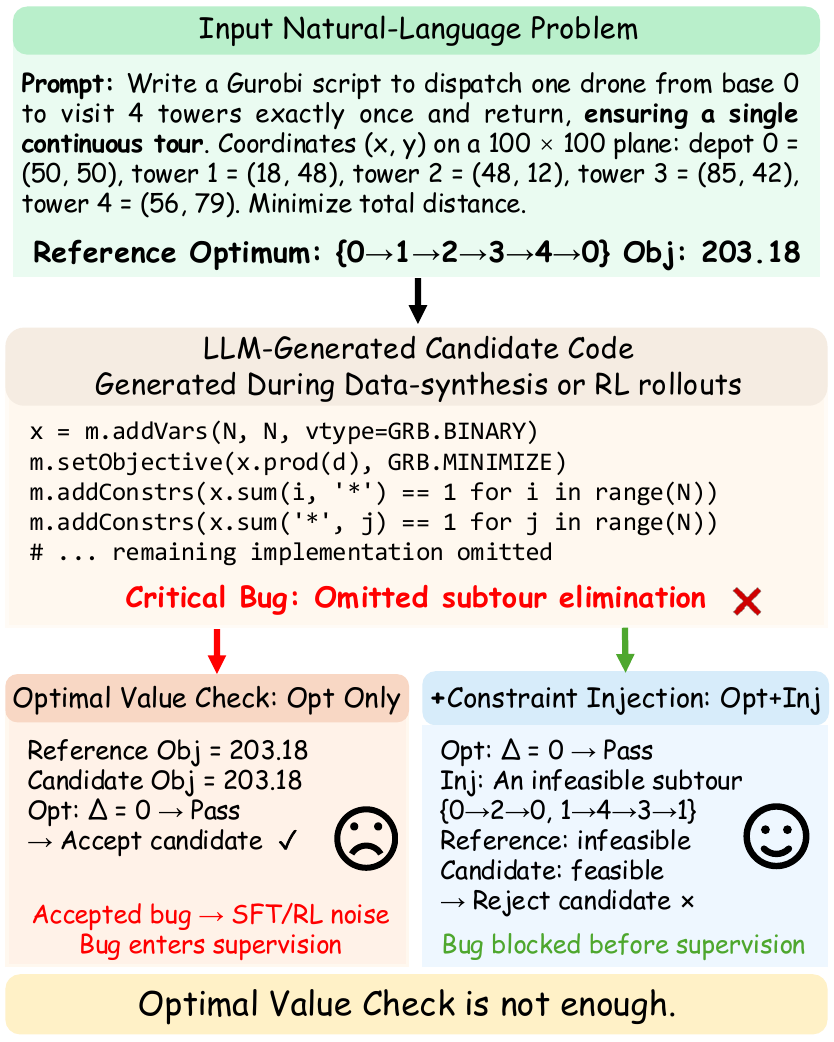}
    \caption{A candidate missing the subtour-elimination constraint still matches the reference optimal value, so the optimal value check accepts it. Constraint injection injects a disconnected-subtour solution: correct code rejects it as infeasible, while the buggy candidate accepts it, exposing the missing constraint.}
    \label{fig:introduction}
\end{figure}

Existing LLM-based optimization modeling methods mainly fall into three categories. Inference-time enhancement methods improve frozen LLMs via prompting, retrieval, self-debugging, or agentic workflows~\citep{jiang2025droc,li2025ars,zhang2025agentic}. Supervised fine-tuning (SFT) trains operations research LLMs (OR-LLMs) on synthetic problem--code pairs, often filtered by executability, optimal value checks, or self consistency~\citep{huang2025orlm,lu2025optmath,zhang2025optimind}. Reinforcement learning (RL) methods use solver feedback such as objective matching or answer agreement~\citep{chen2025sirl,zhou2025steporlm,or2025r1}. Despite their differences, both SFT filtering and RL rewards often reduce to objective equivalence, typically assessed by comparing optimal objective values.

However, objective equivalence does not guarantee constraint fidelity. A generated formulation can match the reference optimum even if it adds redundant constraints or omits required ones, provided that the affected constraints are non-binding on the tested instance. We call these two failure modes \emph{spurious over-constraint} and \emph{silent constraint omission}. Since both can survive executability and optimal value checks, they can contaminate SFT data and be rewarded during RL. Figure~\ref{fig:introduction} shows such a case in a routing instance where the omission of a subtour-elimination constraint leaves the optimum unchanged.

We address this gap with constraint injection, a verification operator that checks whether generated code accepts a feasible solution and rejects one-constraint-violating solutions. Combined with the optimal value check, it forms a dual verifier whose constraint-level signal is decoupled from the optimum on any single instance. We instantiate the dual verifier on vehicle routing problems (VRPs), a constraint-dense testbed whose capacities, time windows, depots, and service rules are coupled into natural-language scenarios~\citep{toth2014vehicle,laporte2009fifty,LAHYANI20151}. The verifier is reused first as a rejection-sampling filter during data synthesis and then as a per-rollout reward during RL. The resulting model, VRPCoder, is an 8B end-to-end model that translates natural-language VRP scenarios into Gurobi code, accompanied by VRPBench\footnote{\url{https://github.com/ccccziiii/VRPBench}}, an expert-verified benchmark covering 21 VRP variants across four subsets.

Our contributions are as follows: (1) We identify objective equivalence as a shared blind spot of SFT data filtering and RL rewards, with two failure modes: spurious over-constraint and silent constraint omission.
(2) We propose constraint injection, a constraint-level verifier that uses feasible solutions to detect spurious over-constraints and one-constraint-violating solutions to detect silent omissions, and combine it with the optimal value check into a dual verifier reused for data synthesis and RL. (3) We introduce VRPCoder together with VRPBench, an expert-verified benchmark covering 21 VRP variants across four subsets. VRPCoder-RL reaches 93.00\% overall Pass@1 on VRPBench, outperforms Gemini-3.1-Pro Preview on three subsets, and substantially outperforms Claude-Sonnet-4.5 and prior OR-LLMs.

\section{Related Work}
\subsection{Supervised Fine-Tuning of OR-LLMs}

Data synthesis has been widely adopted to train OR LLMs~\citep{ijcai2025p1192}. ORLM~\citep{huang2025orlm} expands industry-case seeds with GPT-4; OptMATH~\citep{lu2025optmath} generates complexity-controlled problems and filters them via bidirectional re-modeling; LLMOPT~\citep{jiang2024llmopt} defines a five-element unified formulation for multi-instruction tuning; ReSocratic~\citep{yang2024optibench} derives problem--code pairs from structured demonstrations. Across many such pipelines, sample verification relies mainly on executability, re-modeling consistency, or objective equivalence. While useful for filtering invalid or clearly incorrect code, such signals do not verify whether the generated code faithfully implements the intended constraints, and can therefore miss constraint omissions that leave the optimum unchanged.

\subsection{Reinforcement Learning for OR-LLMs}
Beyond SFT, RL with solver-based verifiable rewards has become a natural direction for OR-LLMs, since generated code can be executed, solved, and evaluated by feasibility, objective values, or answer agreement~\citep{le2022coderl}. SIRL~\citep{chen2025sirl} uses solver execution as outcome feedback, FOARL~\citep{jiang2025foarl} introduces feasibility-and-optimality-aware RL for combinatorial optimization, StepORLM~\citep{zhou2025steporlm} uses a generative process reward model for step-wise supervision, and OR-R1~\citep{or2025r1} performs test-time group relative policy optimization (GRPO) using majority-voting pseudo-labels as the reward. Although these rewards improve OR reasoning beyond SFT, they mainly operate at the solution or answer level and do not directly verify whether each required constraint is implemented in the generated code. For constraint-dense VRP solver-code generation, rewarding constraint-level correctness beyond solver outcomes or answer agreement remains underexplored.

\subsection{LLMs for Vehicle Routing Problems}
VRP is a representative testbed for LLM-based optimization modeling under coupled constraints. NLCO~\citep{jiang2026nlco} benchmarks LLM reasoning on natural-language combinatorial optimization including routing tasks; \citet{huang2024words} show that general-purpose LLMs exhibit preliminary VRP code-generation ability but remain limited under complex constraints. Another line of work improves VRP solving through inference-time mechanisms: DRoC~\citep{jiang2025droc} combines constraint decomposition with retrieval and self-debugging, while ARS~\citep{li2025ars} and AFL~\citep{zhang2025agentic} adopt agentic workflows over frozen LLMs. These studies do not directly address whether generated solver code faithfully encodes intended constraints; this is the focus of our work.

\section{Preliminaries}
\subsection{Problem Definition}
\label{sec:task_definition}

We formulate the task of \emph{automated optimization modeling}: given a natural-language problem description $p$, an LLM generates solver code $y$ that produces an optimal solution and its optimal value upon execution:
\begin{equation}
y = \mathrm{LLM}(p), \quad (s^*, z^*) = \mathrm{Solve}(y)
\label{eq:task}
\end{equation}
where $s^*$ denotes the optimal solution and $z^*$ the corresponding optimal value.
A correct $y$ must (1) produce a feasible optimal solution with the correct optimal value $z^*$ and (2) faithfully encode the explicit constraints in $p$ together with the implicit constraints of the corresponding problem variant. We use \emph{objective equivalence} to denote agreement in the optimal objective value between generated and reference code on the same instance; such agreement alone does not guarantee faithful constraint encoding. In this work, $y$ is instantiated as Gurobi solver code for VRPs.

\subsection{Vehicle Routing Problem}
\label{sec:vehicle_routing_problem}

The Capacitated Vehicle Routing Problem (CVRP)~\citep{braekers2016vehicle} is adopted as the foundation for VRP variants. Consider a graph with customer set $N=\{1,\ldots,n\}$, depot node $0$, node set $V=N\cup\{0\}$, vehicle set $M$, and arc set $A$. Each customer $i$ has demand $d_i$, each vehicle has capacity $Q$, and each arc $(i,j)$ has cost $c_{ij}$. Let $x_{ij}^{k}\in\{0,1\}$ indicate whether vehicle $k$ traverses arc $(i,j)$, and let $v_i\in[1,n]$ denote the position of customer $i$ in its route. 
The CVRP minimizes total routing cost~\eqref{eq:cvrp-objective} subject to: each customer is visited exactly once~\eqref{eq:cvrp-visit}, each vehicle departs at most once~\eqref{eq:cvrp-depart}, flow conservation~\eqref{eq:cvrp-flow}, vehicle capacity~\eqref{eq:cvrp-capacity}, and subtour elimination via Miller--Tucker--Zemlin (MTZ) ordering~\eqref{eq:cvrp-mtz}:
\begin{align}   
  \min \quad & \sum_{k\in M}\sum_{(i,j)\in A} c_{ij}x_{ij}^{k}, \label{eq:cvrp-objective}\\   
  \sum_{k\in M}\sum_{j:(i,j)\in A} x_{ij}^{k} &= 1, \quad \forall i\in N, \label{eq:cvrp-visit}\\   
  \sum_{j:(0,j)\in A} x_{0j}^{k} &\le 1, \quad \forall k\in M, \label{eq:cvrp-depart}\\   
  \sum_{j:(i,j)\in A} x_{ij}^{k} &= \sum_{j:(j,i)\in A} x_{ji}^{k}, \label{eq:cvrp-flow} \\ 
  & \quad \forall i\in V,\ k\in M, \notag \\ 
% \end{align} 
% \begin{align}   
  \sum_{i\in N} d_i  \sum_{j:(i,j)\in A} &  x_{ij}^{k} \le Q, \quad \forall k\in M, \label{eq:cvrp-capacity} \\   
  v_i-v_j+n \sum_{k\in M} & x_{ij}^{k} \le n-1, \label{eq:cvrp-mtz}\\   
  & \quad \forall i,j\in N,\ i\ne j. \notag   
\end{align}

\textbf{VRP variants.}
We extend CVRP with constraint modules such as time windows, pickup-delivery, backhauls, multi-depot, heterogeneous fleets, and open routes, yielding 21 variants in total: 18 used for training and 3 held out for generalization testing. Full mapping is provided in Appendix~\ref{app:variant_constraint_matrix}.

\subsection{Abstract and Concrete Code}
\label{sec:abstract_concrete_code}

To enable scalable data synthesis, we decompose Gurobi solver code into two layers based on the observation that a VRP instance is fully determined by a parameter set (nodes, vehicles, demands, capacities, costs, and variant-specific fields such as time windows or fleet attributes).

\paragraph{Abstract code $C_{\mathrm{abs}}^{v}$.}
For each variant $v$, an OR expert writes a Gurobi code template that defines the decision variables, constraints, and objective, with instance-specific parameters left as inputs. $C_{\mathrm{abs}}^{v}$ serves as the reusable template for variant $v$.
                    
\paragraph{Concrete code $C_{\mathrm{con}}$.} Instantiating $C_{\mathrm{abs}}^{v}$ with a sampled parameter set $\xi \sim \mathcal{P}_v$ yields executable concrete code:
% \begin{equation}
% C_{\mathrm{con}} = C_{\mathrm{abs}}^{v}(\theta), \quad \theta \sim \mathcal{P}_v
% \label{eq:concrete_code}
% \end{equation}
\begin{equation}
C_{\mathrm{con}} = C_{\mathrm{abs}}^{v}(\xi),
\quad \xi\sim\mathcal P_v
\label{eq:concrete_code}
\end{equation}

where $\mathcal P_v$ is the parameter distribution for variant $v$. Since each $\xi$ yields a distinct instance, arbitrarily many $C_{\mathrm{con}}$ can be generated from a single template.

\section{Method}
\subsection{Overview}
\label{sec:method_overview}
Our framework consists of a data synthesis pipeline and a training pipeline. In the data synthesis pipeline, we instantiate expert-written abstract code into concrete code, construct feasible and one-constraint-violating solutions, back-translate the concrete code into natural-language problems, and regenerate solver code with a teacher LLM. Each generated sample is then filtered by the dual verifier, which jointly applies the optimal value check and the constraint injection check. In the training pipeline, accepted samples are first used for SFT; the same dual verifier is then reused as the reward signal during RL with GRPO. The following sections detail each component.

\subsection{Data Synthesis Pipeline}
\label{sec:data_synthesis}

\begin{figure}[t]
    \centering
    \includegraphics[width=\columnwidth]{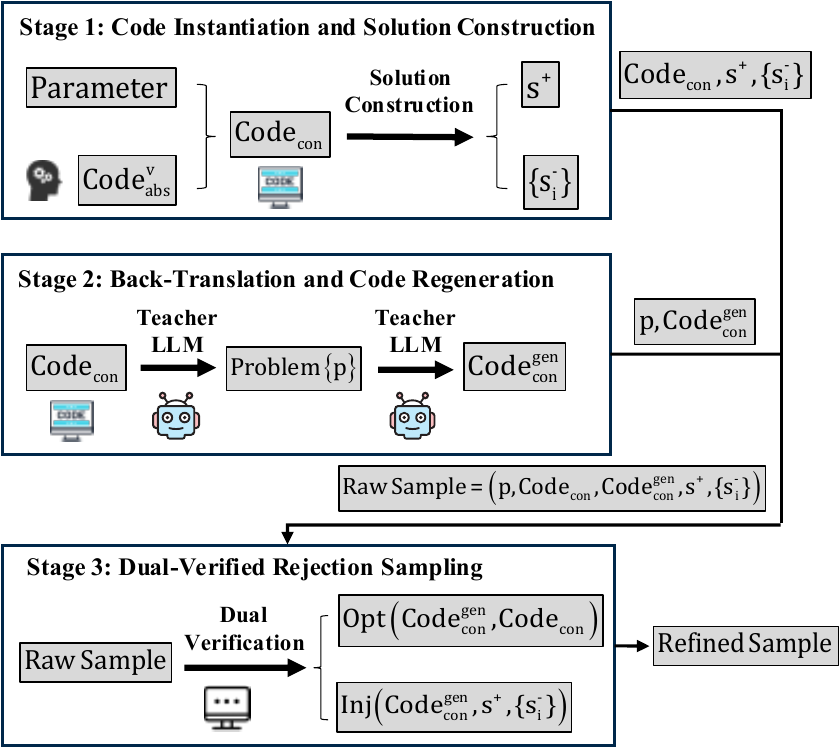}
    \caption{Overview of the three-stage data synthesis pipeline.}
    \label{fig:data_pipeline}
\end{figure}

Figure~\ref{fig:data_pipeline} summarizes our data synthesis pipeline.

\paragraph{Stage 1: Code Instantiation and Solution Construction.}
\label{sec:param_solution}
For each of the 18 training variants, we sample multiple parameter sets and substitute them into $C_{\mathrm{abs}}^{v}$, producing multiple concrete code instances $\{C_{\mathrm{con}}\}$ per abstract template. Specifically, each instance is parameterized by customer count $|N|$, vehicle count $|M|$, customer coordinates, and base demand range. Variant-specific parameters such as capacities, time windows, distance limits, and fleet types are then derived accordingly. We define 12 regular scale profiles and 3 boundary profiles that test extreme geometric layouts, tight capacities, and narrow time-resource budgets. Parameters are randomly sampled or selected under a fixed random seed per instance (see Appendix~\ref{app:profile_design} and Appendix~\ref{app:variant_param_rules}). We then construct a feasible solution $s^+$ and a set of infeasible solutions $\{s_i^-\}$ for each $C_{\mathrm{con}}$ as follows.

\paragraph{Feasible solution $s^+$.}
We construct $s^+$ with a greedy nearest-neighbor heuristic that, starting from a depot, iteratively appends the closest feasible customer to the current route, opens a new vehicle when no extension satisfies capacity or distance limits, and inserts any remaining customers into routes with the largest slack. If the heuristic fails to produce a feasible solution, we replace the objective of $C_{\mathrm{con}}$ with a constant and invoke Gurobi to return any feasible solution under a short time limit.

\paragraph{Infeasible solutions $\{s_i^-\}$.}
For each variant $v$, we define a set of perturbation operators $\mathcal A_v$ that perturb $s^+$ into infeasible solutions $\{s_i^-\}$. 
Each $s_i^-$ violates exactly one class of constraints while satisfying all others, which we term a \emph{one-constraint-violating} solution.
This design isolates the effect of specific categories of constraints, enabling the model to learn fine-grained feasibility reasoning.
The operators fall into two categories depending on whether they modify the parameters of $C_{\mathrm{con}}$.

\begin{figure}[t]
    \centering
    \includegraphics[width=\columnwidth]{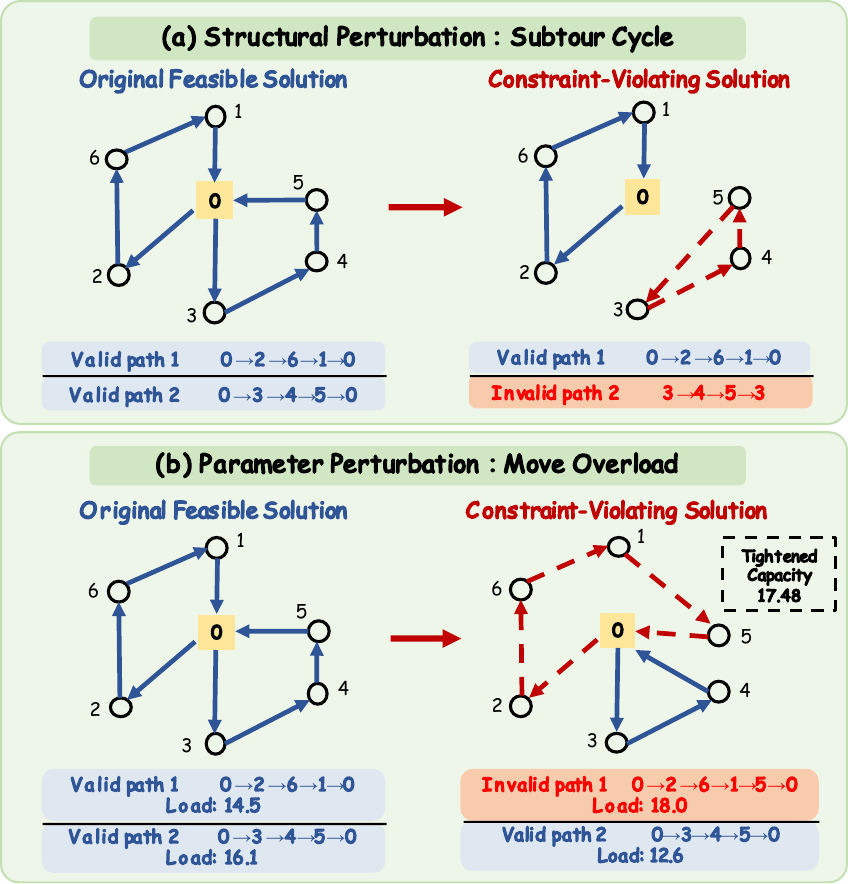}
    \caption{Two illustrative perturbation operators: (a) Subtour
Cycle (structural perturbation); (b) Move Overload (parameter perturbation).}
    \label{fig:attack}
\end{figure}

\begin{itemize}
\item \emph{Structural perturbations} ($\mathcal A_v^{\mathrm{struct}}$) keep all parameters of $C_{\mathrm{con}}$ unchanged and perturb $s^+$ alone, introducing a local structural change that renders the resulting $s_i^-$ infeasible with respect to exactly one class of constraints.
They are applicable to specific categories of constraints---subtour elimination, precedence, and coverage---whose violations are determined by solution topology alone, independent of numerical parameters such as capacities or time windows.
Figure~\ref{fig:attack}(a) illustrates a CVRP example: the \emph{subtour cycle} operator removes the depot connections from route $0\!\to\!3\!\to\!4\!\to\!5\!\to\!0$ to form the closed cycle $3\!\to\!4\!\to\!5\!\to\!3$, which disconnects these nodes from the depot and thereby violates the MTZ subtour elimination constraints (Eq.~\eqref{eq:cvrp-mtz}).

\item \emph{Parameter perturbations} ($\mathcal A_v^{\mathrm{param}}$) construct $s_i^-$ by modifying $s^+$ together with the relevant parameter settings in $C_{\mathrm{con}}$, producing a perturbed concrete code $\tilde C_{\mathrm{con}}$. 
They are needed for resource-bound categories of constraints---capacity, distance, compartment, and hard time windows---where $s^+$ leaves slack under the original parameters, so structural changes alone may not trigger infeasibility. 
The relevant parameter settings are adjusted so that $s^+$ remains feasible under $\tilde C_{\mathrm{con}}$ while $s_i^-$ becomes infeasible.
Figure~\ref{fig:attack}(b) illustrates a CVRP example: the \emph{move overload} operator moves customer~$5$ between routes, changing route loads from $(14.5,16.1)$ to $(18.0,12.6)$, and adjusts capacity to $\tilde Q=17.48$ so that $s^+$ stays feasible while $s_i^-$ violates Eq.~\eqref{eq:cvrp-capacity}.
\end{itemize}

Across the 18 training variants, our operator catalogue contains 5 structural and 7 parameter perturbation operators in total. Per-variant parameter ranges and the full operator catalogue are listed in Appendices~\ref{app:variant_param_rules} and~\ref{app:attack_catalogue}, respectively.

\paragraph{Stage 2: Back-Translation and Code Regeneration.}
\label{sec:problem_gen}

A teacher LLM back-translates each $C_{\mathrm{con}}$ into a set of natural-language problems $\{p\}$ via one core path and two augmentation paths, so that a single $C_{\mathrm{con}}$ yields multiple training samples:
(1)~\emph{Scenario Instantiation} (core path) builds a baseline problem $p_0$ across 60 scenarios via a generate--critique--repair loop~\citep{lu2025optmath};
(2)~\emph{Condensation} (augmentation) eliminates solver-exclusive modeling elements or simplifies business expressions, producing a concise $p_{\mathrm{cond}}$;
(3)~\emph{Index Rewriting} (augmentation) varies node-ID conventions, producing $p_{\mathrm{idx}}$~\citep{jiang2026nlco}.
Subsequently, the same teacher maps each $p$ to a target code $C_{\mathrm{con}}^{\mathrm{gen}}$, which serves as the SFT ground truth. Using $C_{\mathrm{con}}^{\mathrm{gen}}$ rather than $C_{\mathrm{con}}$ as the training target introduces cross-sample diversity in code structure. Accordingly, each sample now takes the form $(p,\,C_{\mathrm{con}},\,C_{\mathrm{con}}^{\mathrm{gen}},\,s^+,\,\{s_i^-\})$. 
Detailed back-translation procedures and prompts are provided in Appendix~\ref{app:synthesis} and Appendix~\ref{app:all_prompts}.

\paragraph{Stage 3: Dual-Verified Rejection Sampling.}
\label{sec:rejection}

Since $C_{\mathrm{con}}^{\mathrm{gen}}$ is generated by the teacher LLM rather than derived from verified templates, it may contain modeling errors. 
To filter out incorrect samples, we apply a dual verification mechanism that checks each $C_{\mathrm{con}}^{\mathrm{gen}}$ with two complementary checks against the ground-truth $C_{\mathrm{con}}$ and the solution set $\mathcal{S}=\{s^+\}\cup\{s_i^-\}_{i=1}^{m}$.

\paragraph{Optimal value check $\mathrm{Opt}$.}
Let $z^*(C)$ denote the optimal value returned by executing code $C$. We independently solve both $C_{\mathrm{con}}^{\mathrm{gen}}$ and the ground-truth $C_{\mathrm{con}}$, then compare their optimal values:
% \begin{equation}
% \mathrm{Opt}(C_{\mathrm{con}}^{\mathrm{gen}}, C_{\mathrm{con}}) \triangleq \bigl|\mathrm{Solve}(C_{\mathrm{con}}^{\mathrm{gen}})-\mathrm{Solve}(C_{\mathrm{con}})\bigr|.
% \end{equation}
\begin{equation}
\mathrm{Opt}(C_{\mathrm{con}}^{\mathrm{gen}}, C_{\mathrm{con}})
\triangleq
\left|z^*(C_{\mathrm{con}}^{\mathrm{gen}})-z^*(C_{\mathrm{con}})\right|.
\end{equation}
The check passes if $\mathrm{Opt}\le\varepsilon_{\mathrm{obj}}$, where $\varepsilon_{\mathrm{obj}}$ is a small numerical tolerance accounting for floating-point differences in solver outputs; if $C_{\mathrm{con}}^{\mathrm{gen}}$ fails to compile or yields an infeasible model, the check fails immediately.
However, matching the optimal value alone cannot guarantee faithful constraint encoding in $C_{\mathrm{con}}^{\mathrm{gen}}$.
This motivates the following constraint injection check.

\paragraph{Constraint injection check $\mathrm{Inj}$.}

Given generated code $C_{\mathrm{con}}^{\mathrm{gen}}$, constraint injection checks whether the encoded constraints are consistent with the labeled solutions $\mathcal{S}$ constructed from $C_{\mathrm{con}}$. Specifically, we do not modify the constraints originally generated by $C_{\mathrm{con}}^{\mathrm{gen}}$, but replace the objective function with a constant and, for each solution $s\in\mathcal{S}$, append arc-level constraints $\Phi(s)$ via Gurobi \texttt{addConstr} calls so that the solver answers a pure feasibility query. 
Let the ground-truth label be $\ell(s)$, where $\ell(s^+)=\mathrm{Feasible}$ and $\ell(s_i^-)=\mathrm{Infeasible}$. 
The constraint injection check passes if and only if:
\begin{equation}
\mathrm{Inj}(C_{\mathrm{con}}^{\mathrm{gen}},s)=\ell(s),\quad \forall\, s\in\mathcal{S}.
\label{eq:inj_pass}
\end{equation}

Recall that $N$ denotes the customer set and $V$ denotes the full node set of the current VRP variant. Let $E(s)\subseteq N\times N$ denote the customer-to-customer arcs used by solution $s$, and $U(s)\subseteq N$ the customers left unserved by $s$. Since a solution contains multiple routes, we further define $E_r(s)$ as the customer-to-customer arcs on route $r$. Below we describe how to construct the injection constraints $\Phi(s)$ from $s$. For each solution $s\in\mathcal S$, we first fix its customer-level routing structure with the following constraint set, denoted by $\Phi_{\mathrm{cust}}(s)$:
\begin{equation}
\begin{gathered}
\sum_{k\in M}x_{uv}^{k}
=
\mathbb{I}[(u,v)\in E(s)],
\qquad \forall\,u,v\in N,\\
\sum_{u\in V\setminus\{c\}}
\sum_{k\in M}x_{uc}^{k}
=
0,
\qquad \forall\,c\in U(s).
\end{gathered}
\label{eq:phi_arc}
\end{equation}
That is, each customer-to-customer arc $(u,v)$ is fixed according to the indicator $\mathbb{I}[(u,v)\in E(s)]$: arcs used in $s$ must be traversed by exactly one vehicle, whereas all customer arcs not appearing in $E(s)$ are deactivated. Moreover, every unserved customer $c\in U(s)$ has its incoming arcs blocked entirely. By injecting $s^+$, we can detect whether $C_{\mathrm{con}}^{\mathrm{gen}}$ introduces spurious constraints that incorrectly exclude a feasible solution. For a homogeneous fleet, this gives $\Phi(s^+)=\Phi_{\mathrm{cust}}(s^+)$.

For an infeasible solution $s_i^-$ under a homogeneous fleet, the customer-level injection $\Phi_{\mathrm{cust}}(s_i^-)$ alone is insufficient: the solver can reassign arcs across vehicles to silently repair the violation. We therefore add route-level consistency constraints:
\begin{multline}
\Phi_{\mathrm{route}}(s_i^-)\!:\;
x_{ij}^{k}=x_{ab}^{k},\\
\forall\, k\in M,\;
\forall\, (i,j),(a,b)\in E_r(s_i^-),\;
\forall\, r.
\label{eq:phi_route}
\end{multline}
This ensures that all customer arcs on the same route $r$ are assigned to a single vehicle, without specifying which one. For a homogeneous fleet, the full injection for $s_i^-$ is thus:
\begin{equation}
\Phi(s_i^-)
=
\Phi_{\mathrm{cust}}(s_i^-)
\cup
\Phi_{\mathrm{route}}(s_i^-).
\label{eq:phi_neg}
\end{equation}
In this way, the violation introduced by perturbation is preserved on its original route, enabling us to check whether $C_{\mathrm{con}}^{\mathrm{gen}}$ misses the corresponding category of constraint. For heterogeneous fleets, both $s^+$ and $\{s_i^-\}$ instead use $x_{uv}^{\kappa_s(u,v)}=1$ for every used customer arc $(u,v)$, where
$\kappa_s(u,v)$ is its assigned vehicle in $s$; unused arcs and unserved
customers are handled as above, with details in Appendix~\ref{app:injection_details}.

\subsection{Model Training Pipeline}
\label{sec:training}

\paragraph{Stage 1: Supervised Fine-Tuning.}

Based on the data synthesis pipeline described above, we construct the verified SFT dataset:
\begin{equation}
\mathcal D=
\bigl\{(p,C_{\mathrm{con}}^{\mathrm{gen}})\bigr\}.
\end{equation}

For the SFT stage, we fine-tune a pre-trained LLM on $\mathcal{D}$, where the input is the problem $p$ and the target is the tokenization $y=(y_1,\ldots,y_T)$ of $C_{\mathrm{con}}^{\mathrm{gen}}$. The training objective minimizes the standard autoregressive negative log-likelihood:
\begin{equation}
\mathcal{L}_{\mathrm{SFT}}(\theta)
= -\mathbb{E}_{(p,y)\sim\mathcal{D}}\left[\sum_{t=1}^{T}\log p_{\theta}(y_t\mid p,y_{<t})\right],
\end{equation}
where $p_{\theta}(y_t\mid p,y_{<t})$ denotes the probability of generating token $y_t$ conditioned on the problem description $p$ and all preceding tokens $y_{<t}$.
The fine-tuned model is denoted \textbf{VRPCoder-SFT}.

\paragraph{Stage 2: Reinforcement Learning with GRPO.}
\label{sec:grpo}

\begin{figure}[t]
    \centering
    \includegraphics[width=\columnwidth]{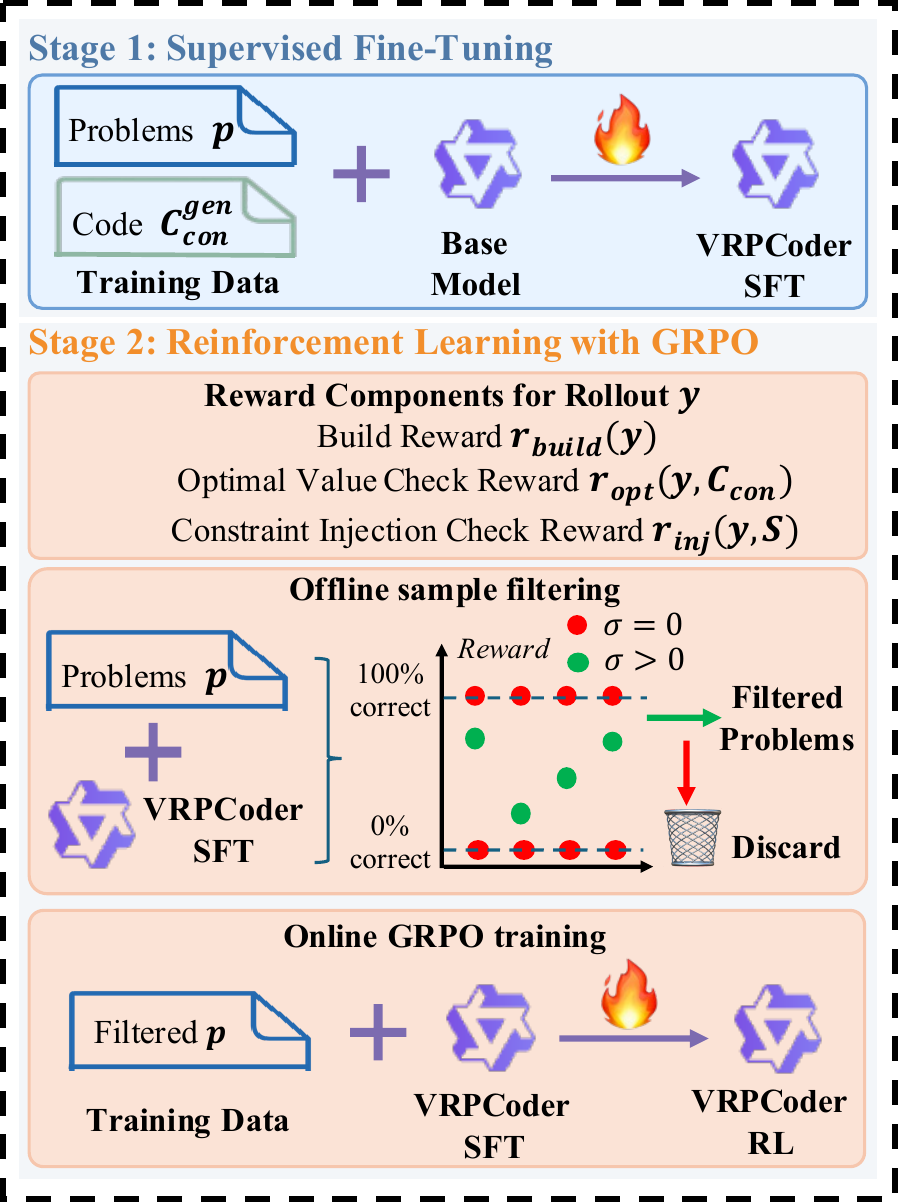}
    \caption{Overview of the model training pipeline.}
    \label{fig:training_pipeline}
\end{figure}

While SFT teaches the model to imitate teacher-generated code, it cannot leverage execution feedback to further improve constraint modeling accuracy. To address this, we apply GRPO~\citep{shao2024deepseekmath} to fine-tune the policy $\pi_\theta$, initialized from VRPCoder-SFT, using reward signals derived from solver execution. Figure~\ref{fig:training_pipeline} summarizes the model training pipeline.
\paragraph{Reward function.}
For a code $y$ generated for problem $p$, the reward is a weighted combination of three components:
\begin{multline}
r(p,y) = \lambda_{\mathrm{build}}\,r_{\mathrm{build}}(y)+\lambda_{\mathrm{opt}}\,r_{\mathrm{opt}}(y,C_{\mathrm{con}})\\
+\lambda_{\mathrm{inj}}\,r_{\mathrm{inj}}(y,\mathcal S),
\end{multline}
where the individual terms are defined as follows: (1) $r_{\mathrm{build}}(y)=\mathbb{I}[\mathrm{build}(y)]$, indicating whether $y$ successfully constructs a valid Gurobi model; (2) $r_{\mathrm{opt}}(y,C_{\mathrm{con}})=\mathbb{I}[\mathrm{Opt}(y,C_{\mathrm{con}})\le\varepsilon_{\mathrm{obj}}]$, indicating whether the optimal value produced by $y$ matches that returned by the reference concrete code $C_{\mathrm{con}}$; and (3) $r_{\mathrm{inj}}(y,\mathcal S)=|\mathcal S|^{-1}\sum_{s\in\mathcal S}\mathbb{I}[\mathrm{Inj}(y,s)=\ell(s)]$, measuring the fraction of solutions in $\mathcal S$ that pass the constraint injection check. Here, $\mathbb{I}[\cdot]$ denotes the indicator function, and $\lambda_{\mathrm{build}}$, $\lambda_{\mathrm{opt}}$, and $\lambda_{\mathrm{inj}}$ denote the weights assigned to the three reward components, respectively.

\paragraph{Offline sample filtering.}
To ensure effective policy optimization, we filter training samples that provide meaningful learning signal. Specifically, for each problem $p$, we pre-generate $H$ rollouts using VRPCoder-SFT and compute the within-group reward standard deviation:
\begin{equation}
\sigma_p = \sqrt{\frac{1}{H}\sum_{h=1}^{H}\bigl(r(p,y_h)-\bar{r}_p\bigr)^2},
\end{equation}
where $\bar{r}_p=\frac{1}{H}\sum_{h=1}^{H}r(p,y_h)$. We retain only samples with $\sigma_p>0$, discarding groups whose rollouts receive identical rewards, as they provide no within-group learning signal.

\paragraph{Online GRPO training.}
On the dataset $\mathcal{D}_{\mathrm{GRPO}}$ obtained after sample filtering, for each problem $p$ we sample $G$ rollouts from $\pi_{\theta_{\mathrm{old}}}$, compute group-normalized advantages, and update $\pi_\theta$ via:

\begin{equation}
\mathcal{J}(\theta)
=\mathbb{E}\!\left[
\frac{1}{G}\sum_{i=1}^{G}\frac{1}{|o_i|}\sum_{t=1}^{|o_i|}
\!\Big(\mathcal{L}^{\mathrm{clip}}_{i,t}-\beta\,D_{\mathrm{KL}}^{(i,t)}\Big)
\right],
\label{eq:grpo_objective}
\end{equation} 

where the expectation is over prompts
$p\sim\mathcal{D}_{\mathrm{GRPO}}$ and rollout groups
$\{o_i\}_{i=1}^{G}$ sampled from
$\pi_{\theta_{\mathrm{old}}}(\cdot\mid p)$.
For token $t$ in rollout $o_i$, the importance ratio is
$\rho_{i,t}
=
\pi_{\theta}(o_{i,t}\mid p,o_{i,<t})
/
\pi_{\theta_{\mathrm{old}}}(o_{i,t}\mid p,o_{i,<t})$.
The clipped surrogate is
$\mathcal{L}^{\mathrm{clip}}_{i,t}
=
\min\!\left(
\rho_{i,t}\hat A_i,
\bar\rho_{i,t}\hat A_i
\right)$,
where
$\bar\rho_{i,t}
=
\mathrm{clip}(\rho_{i,t},1-\epsilon,1+\epsilon)$.
The rollout-level group-normalized advantage is
$\hat A_i=(r(p,o_i)-\mu_r)/(\sigma_r+\varepsilon_{\mathrm{adv}})$, where $\mu_r$ and $\sigma_r$ are the mean and standard deviation of
$\{r(p,o_j)\}_{j=1}^{G}$, and $\varepsilon_{\mathrm{adv}}>0$ is a small
constant for numerical stability; the same $\hat A_i$ is shared across all tokens in $o_i$.
Finally, $D_{\mathrm{KL}}^{(i,t)}$ denotes the token-level KL penalty
against the reference policy $\pi_{\mathrm{ref}}$, a frozen copy of
VRPCoder-SFT.

Here $\beta$ is the KL coefficient, $\epsilon$ is the clipping range,
and $G$ is the number of rollouts per problem. The resulting model is denoted \textbf{VRPCoder-RL}.

\section{Experiments}
\begin{table*}[t]
\centering
\small
\begin{tabular}{lccccc}
\toprule
 & \multicolumn{5}{c}{VRPBench} \\
\cmidrule(lr){2-6}
Model & Subset 1 & Subset 2 & Subset 3  & Subset 4 & AVG \\
\midrule
\multicolumn{6}{l}{\textit{Closed-source frontier LLMs}} \\
Gemini-3.1-Pro Preview & \underline{95.81} & \underline{89.39} & \underline{91.89} & \textbf{96.40} & \textbf{95.00} \\
Claude-Sonnet-4.5 & 70.65 & 42.42 & 79.73 & 58.80 & 64.71 \\
\midrule
\multicolumn{6}{l}{\textit{Open-source general-purpose LLMs}} \\
DeepSeek-V3.2 & 31.29 & 7.58 & 51.35 & 45.60 & 36.29 \\
MiniMax-M2.5 & 21.61 & 9.09 & 17.57 & 32.80 & 24.00 \\
Qwen2.5-72B-Instruct & 15.81 & 6.06 & 12.16 & 27.20 & 18.57 \\
Qwen3-8B & 0.00 & 0.00 & 1.35 & 1.20 & 0.57 \\
\midrule
\multicolumn{6}{l}{\textit{OR-LLMs}} \\
ORLM-LLaMA-3-8B & 3.87 & 0.00 & 12.16 & 7.60 & 5.71 \\
OptMATH-Qwen2.5-7B & 0.97 & 0.00 & 2.70 & 8.40 & 3.71 \\
SIRL-Gurobi-8B & 1.94 & 0.00 & 0.00 & 13.20 & 5.57 \\
SIRL-Gurobi-32B & 8.71 & 3.03 & 1.35 & 30.00 & 15.00 \\
\midrule
\multicolumn{6}{l}{\textit{Our 8B model}} \\
VRPCoder-SFT & 92.26 & \underline{89.39} & 82.43 & 85.20 & 88.43 \\
\textbf{VRPCoder-RL} & \textbf{96.13} & \textbf{96.97} & \textbf{93.24} & \underline{88.00} & \underline{93.00} \\
\bottomrule
\end{tabular}

% \caption{Pass@1 (\%) on VRP solver code generation.}
\caption{Pass@1 (\%) on the four subsets of VRPBench. Best scores are \textbf{bolded}; second-best are \underline{underlined}. AVG is computed over all samples in VRPBench.}
\label{tab:main_results}
\end{table*}

\begin{table*}[t]
\centering
\small
\begin{tabular}{lcccccc}
\toprule
 & \multicolumn{6}{c}{VRPBench} \\
\cmidrule(lr){2-7}
Method & Subset 1 & Subset 2 & Subset 3 & Subset 4 & AVG & $\Delta$ AVG \\
\midrule
VRPCoder-SFT w/o Injection & 91.61 & 81.82 & 83.78 & 79.60 & 85.57 & -- \\
VRPCoder-SFT & 92.26 & 89.39 & 82.43 & 85.20 & 88.43 & +2.86 \\
\midrule
VRPCoder-RL w/o Injection & 94.19 & 93.94 & 91.89 & 80.40 & 89.00 & -- \\
\textbf{VRPCoder-RL} & \textbf{96.13} & \textbf{96.97} & \textbf{93.24} & \textbf{88.00} & \textbf{93.00} & \textbf{+4.00} \\
\bottomrule
\end{tabular}
\caption{Ablation of constraint injection for SFT and RL on VRPBench. Best scores are \textbf{bolded}. $\Delta$ AVG reports the percentage-point gain in overall Pass@1 over the corresponding model without injection.}
\label{tab:ablation_injection}
\end{table*}

\subsection{Experimental Setup}

\paragraph{Benchmark.}
The training dataset $\mathcal{D}$ contains 6{,}797 $(p,\,C_{\mathrm{con}}^{\mathrm{gen}})$ pairs produced by the synthesis pipeline in Section~\ref{sec:data_synthesis}.
For evaluation, we introduce \textbf{VRPBench}, a 700-problem expert-verified benchmark comprising four subsets:
(1)~\emph{In-domain synthetic} (310 problems), generated from the 18 training variants with unseen random seeds, evaluating in-distribution generalization;
(2)~\emph{Out-of-domain synthetic} (66 problems), generated from the 3 held-out compositional variants, evaluating whether the model can combine familiar constraint primitives into unseen VRP variants;
(3)~\emph{Expert-written} (74 problems), sampled from public optimization libraries and formulated by OR experts, evaluating performance on human-authored problem descriptions;
(4)~\emph{Externally-sourced} (250 problems), collected from MAMO~\citep{huang2024mamo}, IndustryOR~\citep{huang2025orlm}, OptMATH~\citep{lu2025optmath}, and NLCO~\citep{jiang2026nlco}, evaluating cross-source robustness.

\paragraph{Baselines.}
We compare against four baseline groups:
(1) closed-source frontier LLMs, including Gemini-3.1-Pro Preview~\citep{googledeepmind2026gemini} and Claude-Sonnet-4.5~\citep{anthropic2025claude};
(2) open-source general-purpose LLMs, including DeepSeek-V3.2~\citep{deepseekai2025deepseekv3}, MiniMax-M2.5~\citep{minimax2025minimax}, Qwen2.5-72B-Instruct~\citep{yang2024qwen2}, and Qwen3-8B~\citep{yang2025qwen3};
(3) SFT-based OR-LLMs, including ORLM-LLaMA-3-8B~\citep{huang2025orlm} and OptMATH-Qwen2.5-7B~\citep{lu2025optmath};
(4) RL-based OR-LLMs, including SIRL-Gurobi-8B and SIRL-Gurobi-32B~\citep{chen2025sirl}. 
We evaluate groups (1) and (2) with the shared code-regeneration prompt (Appendix~\ref{app:all_prompts}), and evaluate groups (3) and (4) with their original prompts and solver settings.

\paragraph{Implementation Details.}
\label{sec:experimental}
Both stages use a LoRA adapter with rank $r=16$, $\alpha=32$, and dropout 0.1 on Qwen3-8B, trained on $4{\times}$A100-40GB GPUs. 
We first apply LoRA SFT on $\mathcal{D}$ for 3 epochs with a maximum sequence length of 8192, using AdamW with learning rate $2\times10^{-4}$ and cosine scheduling, to obtain VRPCoder-SFT. 
We then perform offline sample filtering with $H=6$ pre-generated rollouts per prompt to select frontier prompts $\mathcal{D}_{\mathrm{GRPO}}$, and run GRPO with $G=8$ rollouts per prompt, temperature 0.4, top-$p$ 0.95, and KL coefficient $\beta=0.02$. 
The composite reward weights are $0.2$, $0.5$, and $0.3$ for $r_{\mathrm{build}}$, $r_{\mathrm{opt}}$, and $r_{\mathrm{inj}}$, respectively, yielding VRPCoder-RL. Full settings are provided in Appendix~\ref{app:training_details}.

\subsection{Main Results}
\label{sec:main_results}
Table~\ref{tab:main_results} reports Pass@1 on the four subsets of VRPBench. 
VRPCoder-RL outperforms Gemini-3.1-Pro Preview on Subsets 1, 2, and 3, achieving 96.13\%, 96.97\%, and 93.24\% versus 95.81\%, 89.39\%, and 91.89\%, but trails on Subset 4, which includes the traveling salesman problem with time windows (TSPTW), a variant absent from training. 
It exceeds Claude-Sonnet-4.5 by a large margin, with 93.00\% versus 64.71\% overall Pass@1.
Existing OR-LLMs achieve notably low performance across all subsets, where the strongest, SIRL-Gurobi-32B, reaches only 15.00\% average Pass@1, indicating limited applicability to real-world VRP scenarios and further demonstrating the value of VRPBench as a challenging evaluation benchmark. 
Before fine-tuning, the base model Qwen3-8B is essentially non-functional, achieving only 0.57\% overall Pass@1; after our full pipeline, it achieves 93.00\% overall Pass@1, confirming the effectiveness of the proposed training framework.

\subsection{Ablation Studies}
\label{sec:ablation}

Table~\ref{tab:ablation_injection} isolates the contribution of constraint injection. 
The no-injection setting removes the injection signal from both stages: SFT retains samples that pass the optimal value check but fail injection, and RL scores rollouts using only build success and the optimal value check.
Removing injection lowers overall Pass@1 by 2.86 percentage points for SFT (88.43\% to 85.57\%) and by 4.00 points for RL (93.00\% to 89.00\%). The largest gain appears on Subset 4, where RL improves from 80.40\% to 88.00\%. Subset 2 also improves substantially at the SFT stage, from 81.82\% to 89.39\%.
These results indicate that constraint-level supervision provides information beyond the optimal value check, especially under distribution shift and unseen constraint combinations.

\subsection{Model Analysis}
\label{sec:model}

\begin{figure}[t]
\centering
\includegraphics[width=\linewidth]{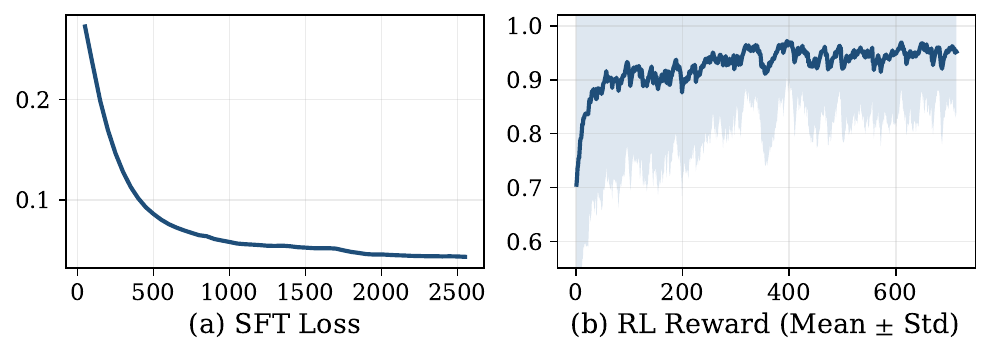}
\caption{Training dynamics of SFT and RL. The left panel shows the SFT training loss, while the right panel shows the RL mean reward. The dark curve indicates the smoothed trend, and the shaded region visualizes local variation over training steps.}
\label{fig:training}
\end{figure}

\paragraph{Training dynamics.}
Figure~\ref{fig:training} shows the training curves of both stages.
In the SFT stage, the training loss decreases steadily from 0.272 to 0.043
over 2{,}550 steps, indicating stable convergence.
In the RL stage, the mean reward shows a clear overall upward trend over
714 steps, with its late-stage level around 0.94.

\paragraph{Sensitivity Analysis.}

\begin{table}[t]
\centering
\small
\setlength{\tabcolsep}{6pt}
\renewcommand{\arraystretch}{0.8}
\begin{tabular}{cccc}
\toprule
$\lambda_{\mathrm{build}}$ & $\lambda_{\mathrm{opt}}$ & $\lambda_{\mathrm{inj}}$ & AVG Pass@1 \\
\midrule
0.2 & 0.6 & 0.2 & 91.57 \\
0.2 & 0.5 & 0.3 & \textbf{93.00} \\
0.2 & 0.4 & 0.4 & 92.86 \\
0.2 & 0.3 & 0.5 & 90.71 \\
\bottomrule
\end{tabular}
\caption{Sensitivity analysis of the three reward weights ($\lambda_{\mathrm{build}}$, $\lambda_{\mathrm{opt}}$, $\lambda_{\mathrm{inj}}$) for VRPCoder-RL. Best score is \textbf{bolded}.}
\label{tab:reward_weights}
\end{table}

Table~\ref{tab:reward_weights} shows the reward weight sensitivity analysis. We keep the three weights summing to 1, fix $\lambda_{\mathrm{build}}{=}0.2$, and vary $\lambda_{\mathrm{opt}}$ and $\lambda_{\mathrm{inj}}$. Pass@1 remains above 90\% across all configurations and peaks at 93.00\% for $\lambda_{\mathrm{opt}}/\lambda_{\mathrm{inj}}=0.5/0.3$. It is slightly lower when injection is down-weighted (0.6/0.2: 91.57\%) or made dominant (0.3/0.5: 90.71\%).
% The optimal value check is a dense objective comparison while constraint injection is a sparse constraint-level signal, so constraint injection works best as a complement rather than the primary reward. 
This suggests that constraint injection works best as a complement to the optimal value check rather than as the dominant reward.
All weight configurations outperform removing injection entirely (Table~\ref{tab:ablation_injection}).

\section{Conclusion}
We identified objective equivalence as a shared blind spot in SFT- and RL-based optimization-modeling pipelines, where spurious over-constraints and silent constraint omissions can be indistinguishable from correct code when the affected constraints are non-binding at the optimum. We proposed \textbf{constraint injection}, a constraint-level verification operator that uses feasible and one-constraint-violating solutions; combined with the optimal value check, it forms a dual verifier. We applied this verifier as a rejection-sampling filter during data synthesis and a per-rollout reward in GRPO. When instantiated on vehicle routing as a constraint-dense testbed, our framework yields VRPCoder-RL, which reaches 93.00\% overall Pass@1 on VRPBench. It is competitive with frontier LLMs and surpasses prior OR-LLMs by a large margin. Ablations show that constraint injection is a key component: the largest improvement appears on the most distribution-shifted subset, and constraint-level supervision provides benefits beyond the optimal value check.

\section*{Limitations}
Our empirical validation focuses on vehicle routing, and transferring constraint injection to other optimization domains requires expert-written abstract-code templates, constraint-family-specific perturbation operators, and compatible solution-encoding rules. This effort is incurred once per recurring constraint family rather than per instance, but transferability beyond routing remains to be evaluated. Our one-constraint-violating solutions isolate individual constraint families while the other constraints in the generated model remain active, but they may not expose errors that arise only through interactions among multiple constraints. We report instance-level Pass@1 to enable consistent comparison across all four subsets of VRPBench. Although Appendix~\ref{app:constraint_error_analysis} reports constraint-level failures during data synthesis, benchmark-wide constraint-level evaluation of final model outputs remains future work.

\section*{Acknowledgments}
This research was supported by the National Natural Science Foundation of China under Grant 72571017.
\bibliography{custom}
\newpage
\appendix
\section{VRP Variant Overview}
\label{app:variants_and_attacks}
\label{app:variant_constraint_matrix}

We build 21 VRP variants. The 21-variant count refers to the variants constructed for Subsets 1--3; the externally sourced Subset 4 additionally includes TSPTW.

\textbf{MD (multi-depot)} introduces multiple depots, with each vehicle departing from and returning to its assigned depot. \textbf{MC (multi-compartment)} partitions each vehicle into multiple non-mixing compartments, each with an independent capacity. \textbf{PD (pickup-and-delivery)} pairs pickup and delivery nodes that must be visited by the same vehicle in order. \textbf{Open (open routing)} allows each route to terminate at the last customer without returning to the depot. \textbf{HTW (hard time window)} imposes hard time windows $[a_i,b_i]$ on service start times; early arrival is allowed through waiting, but service cannot start before $a_i$ or after $b_i$. \textbf{STW (soft time window)} relaxes time-window violations into objective penalties. \textbf{Dist (distance limit)} imposes a maximum travel distance $D_{\max}$ on each route. \textbf{BH (backhaul)} splits customers into linehaul and backhaul segments and requires all linehaul customers to precede backhaul customers on each route. \textbf{Split (split delivery)} allows a customer to be served by multiple vehicles. \textbf{HF (heterogeneous fleet)} allows capacities, fixed costs, and per-distance costs to vary by vehicle.

The Cap. column in Table~\ref{tab:variant_constraint_matrix} specifies capacity semantics: $Q$ for homogeneous capacity, $Q_k$ for heterogeneous capacity, and ``--'' for no effective capacity limit. VRPB accounts for linehaul and backhaul loads separately, whereas MCVRP and HFMCVRP account for capacities separately for each compartment.

\begin{table*}[t]
\centering
\small
\begin{tabular}{lcccccccccccc}
\toprule
Variant & Cap. & MD & MC & PD & Open & HTW & STW & Dist & BH & Split & HF & Train \\
\midrule
TSP & -- & & & & & & & & & & & \checkmark \\
CVRP & $Q$ & & & & & & & & & & & \checkmark \\
MDVRP & $Q$ & \checkmark & & & & & & & & & & \checkmark \\
MCVRP & $Q$ & & \checkmark & & & & & & & & & \checkmark \\
PDP & $Q$ & & & \checkmark & & & & & & & & \checkmark \\
OVRP & $Q$ & & & & \checkmark & & & & & & & \checkmark \\
DCVRP & $Q$ & & & & & & & \checkmark & & & & \checkmark \\
VRPB & $Q$ & & & & & & & & \checkmark & & & \checkmark \\
SDVRP & $Q$ & & & & & & & & & \checkmark & & \checkmark \\
VRPHTW & $Q$ & & & & & \checkmark & & & & & & \checkmark \\
VRPSTW & $Q$ & & & & & & \checkmark & & & & & \checkmark \\
HFVRP & $Q_k$ & & & & & & & & & & \checkmark & \checkmark \\
HFMDVRP & $Q_k$ & \checkmark & & & & & & & & & \checkmark & \checkmark \\
HFMCVRP & $Q_k$ & & \checkmark & & & & & & & & \checkmark & \checkmark \\
HFVRPHTW & $Q_k$ & & & & & \checkmark & & & & & \checkmark & \checkmark \\
MDVRPSTW & $Q$ & \checkmark & & & & & \checkmark & & & & & \checkmark \\
PDPSTW & $Q$ & & & \checkmark & & & \checkmark & & & & & \checkmark \\
PDPTW & -- & & & \checkmark & & \checkmark & & & & & & \checkmark \\
\midrule
\textbf{\textit{OVRPHTW}} & $Q$ & & & & \checkmark & \checkmark & & & & & & Held-out \\
\textbf{\textit{MCVRPSTW}} & $Q$ & & \checkmark & & & & \checkmark & & & & & Held-out \\
\textbf{\textit{DCVRPHTW}} & $Q$ & & & & & \checkmark & & \checkmark & & & & Held-out \\
\bottomrule
\end{tabular}
\caption{Structural module matrix for the 21 VRP variants. A check mark (\checkmark) indicates the module is enabled. The \textit{Train} column specifies the participation in SFT/GRPO training. \textit{Bold and italicized} variants are strictly held out from the training mixture: their constituent modules are present in the training set, but their pairwise combinations are never seen, forming Subset 2 for evaluating compositional generalization.}
\label{tab:variant_constraint_matrix}
\end{table*}

\section{Constraint-Level Solution Construction}
\label{app:instance_sampling}

In the three-stage data synthesis pipeline (Section~\ref{sec:data_synthesis}), Stage 1 constructs feasible and one-constraint-violating solutions with auditable feasibility labels for each training sample. Given a variant $v$ and a profile, we first determine the customer count, fleet size, coordinate distribution, and base demand range (Appendix~\ref{app:profile_design}). Variant-specific rules then specify capacities, time windows, route-distance limits, heterogeneous vehicle types, pickup--delivery pairs, compartments, and backhaul labels, yielding a complete instance and its reference concrete code $C_{\mathrm{con}}$ (Appendix~\ref{app:variant_param_rules}). We then construct a feasible solution $s^+$ (Appendix~\ref{app:positive_probe}) and derive one-constraint-violating solutions $\{s_i^-\}$ from $s^+$ (Appendix~\ref{app:attack_catalogue}). Finally, bounded resources are slightly relaxed and each labeled solution set $\mathcal S$ is checked by $C_{\mathrm{con}}$ before assembly (Appendix~\ref{app:relaxation_assembly}). The main depot is fixed at $(50,50)$; multi-depot variants use two depots at $(25,50)$ and $(75,50)$; all customer coordinates are clipped to the coordinate range $[5,95]^2$.

\subsection{Profiles}
\label{app:profile_design}

A profile serves as a parameter envelope rather than a training split or an algorithmic component. It is applied before instance generation and specifies the customer count $n$, vehicle count $k$, spatial distribution, and base range for scalar customer demands, while variant-specific modules fill in capacities, time windows, pickup-delivery demand, compartment demand, and other constraints. Profiles are designed so that small-scale instances, regular layouts, and geometric, capacity, or time-boundary cases all lie in a single auditable parameter space, instead of being listed instance by instance. We use 15 profiles in total: 12 regular profiles and 3 boundary profiles; this appendix specifies their parameter coverage. Unless otherwise stated, an interval $[a,b]$ denotes a sampling envelope, and a discrete set lists the admissible choices.

We use $d$ for scalar customer demand, $Q$ for homogeneous vehicle capacity or a capacity bound specified by a boundary-profile grid, $H$ for the time horizon in time-window variants, and $C$ for the number of compartments. A grid tuple such as $(n,k,Q,d)$, $(n,k,Q,H)$, or $(n,k,Q,C)$ denotes an instance-level configuration. If an entry contains an interval, such as $d=[3,6]$, the tuple is first selected and each customer demand is then sampled from that interval. In Table~\ref{tab:profile_definitions}, ``locked'' means that the profile family directly uses the resource values from its grid rather than the default rules in Appendix~\ref{app:variant_param_rules}; ``--'' means that the profile specifies only scale and geometry, and the resource parameters are supplied by the variant-specific rules.

\textbf{Uniform}, \textbf{Clustered}, and \textbf{Radial} distributions are inspired by the VRP instance designs of \citet{augerat1995computational} and \citet{uchoa2017new}; time-window generation follows \citet{solomon1987algorithms}; and regular geometric arrays follow \citet{golden1998impact}. These references serve only as sources of design factors and do not indicate that our samples are drawn from the corresponding benchmarks.

The five spatial distributions define geometric families for customer coordinates only, rather than complete VRP instances. Capacities, time windows, pickup-delivery pairs, compartments, and other constraint modules are supplied by Appendix~\ref{app:variant_param_rules}. \textbf{Uniform} samples customer coordinates independently in the plane. \textbf{Clustered} first samples 2--3 centers and then draws compact customer groups around them. \textbf{Radial} samples angles and radii around the depot. \textbf{Elongated} forms a narrow corridor along a random direction. \textbf{Split} places customers in two separated groups.

\begin{table*}[t]
\centering
\small
\begin{tabular}{rllclll}
\toprule
\# & Profile & $n$ & $k$ & Spatial distribution & Demand $d$ & $Q$ \\
\midrule
1 & Tiny uniform & 5 & 2 & Uniform & $[2,10]$ & -- \\
2 & Tiny clustered & 6 & 2 & Clustered & $[1,10]$ & -- \\
3 & Small radial & 7 & 2 & Radial & $[2,8]$ & -- \\
4 & Small elongated & 7 & 3 & Elongated & $[1,10]$ & -- \\
5 & Medium uniform & 8 & 3 & Uniform & $[1,10]$ & -- \\
6 & Medium clustered & 9 & 3 & Clustered & $[2,12]$ & -- \\
7 & Medium split & 10 & 3 & Split & $[1,8]$ & -- \\
8 & Medium light & 8 & 3 & Uniform & $[0.5,3]$ & -- \\
9 & Medium heavy & 9 & 4 & Uniform & $[5,15]$ & -- \\
10 & Large uniform & 12 & 4 & Uniform & $[1,8]$ & -- \\
11 & Large clustered & 12 & 4 & Clustered & $[2,10]$ & -- \\
12 & Large mixed & 12 & 4 & Elongated & $[1,12]$ & -- \\
13 & Geometry boundary & 4--8 & 1--4 & Deterministic geometry & Family-specific & Partially locked \\
14 & Capacity boundary & 5--12 & 1--4 & Uniform / split clusters & Family-specific & Locked \\
15 & Time / multi-resource boundary & 4--12 & 1--4 & Radial / split clusters & Family-specific & Locked \\
\bottomrule
\end{tabular}
\caption{Compact profile definitions. The demand column gives the default scalar demand range. Zero scalar demand for TSP, pickup-delivery demand, and SDVRP split-delivery demand are supplied by the variant-specific rules. The $Q$ column indicates only whether capacity-related values are fixed at the profile level; the time horizon $H$ and compartment count $C$ are specified only in the family descriptions for profile 15.}
\label{tab:profile_definitions}
\end{table*}

For profile 13, only the hexagon subfamily fixes $Q=100$, keeping capacity safely non-binding so that the resulting solutions focus on depot connectivity and subtour structure; other geometry-boundary subfamilies still take capacity values from Appendix~\ref{app:variant_param_rules}. Profiles 14 and 15 are marked as locked because they are generated by directly sampling grid tuples such as $(n,k,Q,d)$, $(n,k,Q,H)$, or $(n,k,Q,C)$.

The boundary profiles specify parameter arrangements rather than new VRP variants or training subsets. Profile 13 covers deterministic geometric structures. Its hexagon subfamily uses $n=6,k=1$, places customers on a regular hexagon with radius $r\in[15,25]$, places the depot at the midpoint of one hexagon edge, and fixes $Q=100$. The polygon subfamily samples $n\in\{4,6,8\}$ and places customers on a regular polygon centered at $c\in[45,55]^2$, with radius $r\in[25,40]$ and a random rotation. The collinear / coverage subfamily places $n\in\{5,6,7,8\}$ customers at equal spacing on a randomly rotated segment, with spacing in $[3,8]$. The local-cluster subfamily places customers within radius $r\in[1,3]$ around a center in $[20,80]^2$.

Profile 14 samples from several $(n,k,Q,d)$ grids so that the average route demand $\bar d n/k$ is close to the capacity bound. The small-grid settings include $(5,2,12,[3,6])$, $(6,2,14,[3,7])$, $(6,3,12,[2,5])$, $(8,3,16,[2,6])$, $(6,3,8,[3,4])$, $(8,4,8,[3,4])$, $(6,2,12,[3,5])$, and $(8,3,12,[3,5])$. To cover split delivery and heavy-route cases, the extended grids allow $n$ up to 12, $k$ up to 4, scalar demand up to $[2,20]$, and capacities within the envelope specified in Appendix~\ref{app:variant_param_rules}. Customer coordinates use uniform or split-cluster layouts.

Profile 15 includes time-window, pickup-delivery, and compartment subfamilies. The time-window subfamily arranges customers in a radial sequence, uses $d\in[1,4]$, and samples $(n,k,Q,H)$ from four configurations: $(8,3,60,240)$, $(10,3,70,260)$, $(10,4,60,260)$, and $(12,4,80,300)$. The pickup-delivery time-window subfamily uses one vehicle and a longer time horizon, with $n\in\{4,6,8,10,12\}$; pairing is supplied by the variant rules. The compartment subfamily uses a split-cluster layout and samples $(n,k,Q,C)$ from four configurations: $(8,3,24,2)$, $(10,3,26,2)$, $(10,4,24,2)$, and $(12,4,28,2)$.

\subsection{Variant-Specific Parameters}
\label{app:variant_param_rules}

After a profile determines $(n,k)$, the spatial distribution, and the base demand, the variant generator generates the additional parameters required by each constraint module. Unless otherwise specified, continuous values are sampled within the reported intervals and discrete values are selected from the reported sets.

\paragraph{Capacity.}
For homogeneous-capacity variants, effective vehicle capacities lie within the envelope $[8,100]$. The capacity-boundary family does not sample from the default envelope; it directly uses the $(n,k,Q,d)$ grid from Appendix~\ref{app:profile_design}, making the average route demand close to the capacity bound. TSP has no effective capacity constraint and assigns zero demand to nodes. SDVRP uses split-delivery demand, with scalar demand in $[2,20]$ and vehicle capacity in $[30,85]$. MCVRP, HFMCVRP, and MCVRPSTW use two compartments. Each customer has per-compartment demand sampled from $[0,6]$; per-compartment capacity is obtained by splitting vehicle capacity across compartments and applying a mild perturbation, yielding values within the range $[5,55]$. The total vehicle capacity equals the sum of compartment capacities.

\paragraph{Heterogeneous fleets.}
HFVRP, HFVRPHTW, HFMDVRP, and HFMCVRP use three vehicle types. Type-specific capacities, fixed costs, and per-distance costs fall within $[10,65]$, $[20,70]$, and $[0.8,1.2]$, respectively. This covers small, medium, and large vehicles, requiring the model to index both capacity and cost by vehicle type rather than reducing the instance to a homogeneous fleet. In HFMCVRP, the three total base capacities are $\{15,30,50\}$ and then split into two compartments within each type.

\paragraph{Route-distance limits.}
DCVRP and DCVRPHTW use route-level distance limits $D_{\max}$ within the range $[60,1000]$. When a distance-boundary perturbation is constructed, $D_{\max}$ is tightened by the operator according to the route lengths of $s^+$ and the corresponding solution $s_i^-$.

\paragraph{Time windows.}
Regular time-window variants use $H=500$, except PDPTW, which uses a chain-length-dependent horizon $H=\max(1200,\,120p+600)$, where $p$ is the number of pickup-delivery
pairs, yielding $H=1200$ for $p\le 5$ and $H=1320$ for $p=6$. After resource relaxation, customer window endpoints remain within $[0,1500]$. Depot windows are $[0,H]$. Customer service times lie within the effective envelope $[0,10]$: PDPTW uses zero service time, while the remaining time-window variants sample positive values, with DCVRPHTW allowing values up to \(10\). Let $d_0$ be the Euclidean distance from a customer to the main depot. Customer window starts are sampled from intervals proportional to $d_0$, lie in $[0,250]$, and have widths in $[30,120]$. For MDVRPSTW, $d_0$ is instead the distance to the nearest depot when generating each customer's soft time window; this does not assign the customer to that depot. During initial instance generation, PDPSTW and PDPTW first sample pickup windows and then derive the paired delivery windows. PDPSTW initializes each delivery-window start as the pickup start plus pickup service and \(0.5\) times the direct pickup--delivery distance, whereas PDPTW uses the pickup start plus \(0.3\) times that distance with zero service time. These windows may subsequently be widened during feasibility-preserving relaxation. The time-boundary family directly uses the $(n,k,Q,H)$ grid in Appendix~\ref{app:profile_design}.

\paragraph{Depots, pickup-delivery pairs, backhauls, and open routes.}
MDVRP, MDVRPSTW, and HFMDVRP use two depots. PDP and PDPSTW set scalar node demand to zero and sample paired pickup--delivery demand from $[1,8]$, whereas PDPTW uses unit demand for every pair. All three construct pickup--delivery pairs; their customer counts are even and lie in $\{4,6,8,10,12\}$. VRPB labels customers as linehaul or backhaul and requires linehaul-first service. OVRP and OVRPHTW allow routes to terminate at the last customer without counting a return-to-depot leg.

\subsection{Feasible Solution Construction}
\label{app:positive_probe}

For each complete instance, we first run a variant-specific heuristic to obtain $s^+$. The heuristic is mainly based on nearest-neighbor insertion and checks common VRP resource constraints, such as capacity, route-distance limits, and time windows, at each step. If customers remain uninserted, they are preferentially inserted into the route with the largest remaining slack. If the heuristic fails, we invoke the reference concrete code $C_{\mathrm{con}}$ for a time-limited feasibility solve with \texttt{TimeLimit = 15} seconds and \texttt{MIPFocus = 1}. The returned solution is then checked by $C_{\mathrm{con}}$ through a feasibility assertion. If any assertion fails, the candidate sample is discarded. The resulting $s^+$ is used to construct the one-constraint-violating solutions $\{s_i^-\}$.

\subsection{One-Constraint-Violating Solution Construction}
\label{app:attack_catalogue}

Each variant $v$ is associated with a perturbation-operator set $\mathcal A_v=\mathcal A_v^{\mathrm{struct}}\cup\mathcal A_v^{\mathrm{param}}$. Each operator transforms $s^+$ into a one-constraint-violating solution $s_i^-$ that targets one active constraint family. Structural perturbations change only the customer visiting sequence and do not modify instance parameters. Parameter perturbations change the visiting sequence and tighten one family of resource bounds. We construct these solutions for logical structure constraints such as customer coverage, depot connectivity, and pickup--delivery precedence, and for common resource constraints such as capacity, distance, compartments, and hard time windows. Constraints that mainly affect objective penalties rather than producing deterministic infeasibility, such as soft time-window penalties, are handled by the optimal value check.

The perturbation operators only generate candidate solutions. The final labels are assigned by checking each candidate instance--solution pair with $C_{\mathrm{con}}$. Each $s_i^-$ is derived from $s^+$ by one operator and targets one constraint family by construction. If $s^+$ is infeasible on the corresponding instance, or if any $s_i^-$ is not rejected by $C_{\mathrm{con}}$, the candidate is discarded.

Each $s_i^-$ is injected separately into the complete generated model, while all other generated constraints remain active. Combining several violations in one solution would make the verdict non-diagnostic, because the model could reject that solution by correctly enforcing any one of the violated constraints while still omitting another. We therefore require $C_{\mathrm{con}}^{\mathrm{gen}}$ to accept $s^+$ and reject every $s_i^-$ separately. This design isolates individual constraint-family errors but does not exhaustively test errors that emerge only through interactions among multiple constraints.

SDVRP is handled separately because split delivery changes the semantics of coverage, subtours, and capacity. A customer can be served by multiple vehicles; removing a customer must remove all of its delivered portions; a subtour customer may otherwise be reconnected to a depot through another route; and overload can be repaired by redistributing demand across vehicles. Therefore, SDVRP uses split-delivery versions of the remove-customer and isolated-subtour perturbations, together with two capacity-oriented perturbations: quarantine overload and phantom delivery. Repeatedly visiting the same customer is not necessarily a violation under split delivery and is therefore not used to construct an $s_i^-$.

\subsubsection{Structural Perturbations}

\paragraph{Remove customer} targets coverage or demand-satisfaction constraints. For standard non-split variants, every customer $c\in N$ must be visited exactly once, e.g., $\sum_{i,k}x_{i,c,k}=1$. The operator selects a visited customer from $s^+$ and removes it from its route; for example, $[0,1,2,0]$ becomes $[0,1,0]$, leaving customer 2 unserved. In SDVRP, where a customer may be split across vehicles, the selected customer is removed from every route that serves any portion of it. If $C_{\mathrm{con}}^{\mathrm{gen}}$ omits the coverage/demand-satisfaction constraint, relaxes it to an inequality, or covers only a subset of customers, the resulting $s_i^-$ may be accepted.

\paragraph{Subtour cycle} targets depot connectivity constraints: every route must be connected through a depot and cannot contain a customer-only cycle. The operator extracts at least two customers from a feasible route and forms a cycle that does not pass through the depot, e.g., a detached cycle such as $[1,2,1]$. If $C_{\mathrm{con}}^{\mathrm{gen}}$ omits subtour elimination or only enforces local flow balance, the detached cycle may be accepted. SDVRP uses an isolated-subtour solution, which additionally removes the cycle customers from other routes so that split service cannot reconnect them to the depot.

\paragraph{Backhaul order violation} targets the VRPB linehaul-first rule: all linehaul customers must precede all backhaul customers on each route, equivalently forbidding any transition from a backhaul customer to a linehaul customer. The operator moves a backhaul customer before a linehaul customer, e.g., changing $[0,l_1,l_2,b_1,0]$ into $[0,l_1,b_1,l_2,0]$. The optimal value check alone cannot verify that this rule is encoded, because the optimum in an enlarged feasible region may not use a reversed edge.

\paragraph{Precedence violation} targets pickup-delivery precedence. All pickup--delivery variants require each pickup to precede its paired delivery on the same route. PDPTW enforces this through service-start times, whereas PDP and PDPSTW use route-order variables. The operator swaps the pickup and delivery within the same route, e.g., $0,p,d,0$ becomes $0,d,p,0$.

\paragraph{Split pickup-delivery pair} targets same-vehicle coupling for pickup-delivery pairs: each pair $(p,d)$ must be served by the same vehicle, expressible as $\sum_j x_{p,j,k}=\sum_j x_{d,j,k},\ \forall k$. The operator keeps the pickup on the original vehicle and moves the delivery into another vehicle route. It exposes models that encode precedence but omit same-vehicle coupling.

\subsubsection{Parameter Perturbations}

Parameter perturbations construct a boundary case for one resource. Let $g$ be the resource usage of the target route, vehicle, or compartment, and let $b$ be the corresponding resource usage of a perturbed solution $s_i^-$, with $b>g$. Let $\tilde B$ be the tightened resource bound in the perturbed instance, such as capacity $\tilde Q$, route-distance limit $\tilde D_{\max}$, or compartment bound $\tilde Q_c$. If $\tilde B\le g$, the feasible solution is incorrectly excluded; if $\tilde B\ge b$, the one-constraint-violating solution no longer violates the resource. Thus, except for time-window violations, parameter perturbations set $\tilde B$ inside the open interval $(g,b)$:
\begin{equation}
\tilde B=\max\!\bigl(g+\varepsilon,\ \min\!\bigl(g+0.85(b-g),\ b-\varepsilon\bigr)\bigr).
\label{eq:param_attack_bound}
\end{equation}
Here $\varepsilon>0$ is a small numerical margin ensuring that $\tilde B$ is strictly between $g$ and $b$. The 0.85 interpolation places the one-constraint-violating solution close to the boundary, testing whether the regenerated model truly encodes that upper bound. Clipping prevents numerical rounding from moving $\tilde B$ onto $g$ or $b$. SDVRP uses a more conservative margin because split-delivery quantities are more sensitive to rounding. The time-window violation operator does not use this scalar bound-tightening rule, because it swaps the visiting order to create hard time-window lateness.

\paragraph{Capacity overload} targets capacity constraints $\sum_i d_i\sum_j x_{i,j,k}\le Q$ for all vehicles $k$. It moves one or more customers from other routes into a target route until the target route exceeds the tightened capacity. We set $g$ to the target route's original load under $s^+$ and $b$ to the load of the overloaded target route. We then tighten $Q$, or the target vehicle's capacity, to a value in $(g,b)$. For homogeneous fleets, the tightened bound is additionally kept above the largest route load in $s^+$; for vehicle-specific bounds, it is applied only to the target vehicle. The resulting $s_i^-$ checks whether capacity is present and applied per vehicle. VRPB computes linehaul and backhaul loads separately and uses their maximum. PDP and PDPSTW move whole pickup-delivery pairs, and rebuild the target route in pickup-before-delivery order, to avoid confounding capacity with same-vehicle or precedence violations.

\paragraph{Heterogeneous fleet overload} targets vehicle-specific capacity constraints, where each vehicle has its own capacity and cost parameters. For HFVRP and HFVRPHTW, we construct an extreme-overload solution: customers are greedily accumulated into a single route until their total demand exceeds the maximum capacity of any vehicle, making the route infeasible under any reassignment. If this construction fails, the implementation falls back to the generic move-overload perturbation. For heterogeneous multi-depot or multi-compartment variants, the generic overload operator tightens only the target vehicle's capacity, while compartment-specific variants additionally use compartment-overload solutions. These perturbations test whether $C_{\mathrm{con}}^{\mathrm{gen}}$ respects vehicle-specific capacities rather than replacing them with a single homogeneous bound.

\paragraph{Distance overrun} targets route-distance limits $\sum_{i,j}c_{i,j}x_{i,j,k}\le D_{\max}$ for all vehicles $k$. It moves customers into the longest route in $s^+$ so that the perturbed route length $b$ exceeds the largest route length in $s^+$, denoted by $g$, and then tightens $D_{\max}$ to a value in $(g,b)$. It tests whether the distance limit is enforced separately for each vehicle and computed from all traveled legs of its route.

\paragraph{Compartment overload} targets compartment capacity constraints $\sum_i d_{i,c}\sum_j x_{i,j,k}\le Q_c$ for each vehicle $k$ and compartment $c$. Multi-compartment routing is not captured by total vehicle load: one compartment can overflow while the total load still appears feasible. The operator concentrates demand from one compartment on a vehicle and tightens only that compartment bound while keeping total vehicle capacity non-binding.

\paragraph{Quarantine overload} targets the SDVRP per-vehicle capacity constraint $\sum_i d_i^k\le Q$ for every vehicle $k$, where $d_i^k$ is the portion of customer $i$ served by vehicle $k$. Ordinary capacity perturbations can be repaired in SDVRP by redistributing demand. This operator fixes the service assignment of the target customers so the diagnostic question becomes whether the model still enforces the capacity bound for that vehicle.

\paragraph{Phantom delivery} targets the coupling between vehicle visits and delivered quantities in SDVRP. Split delivery models commonly use delivery-allocation variables to indicate how much demand of customer $i$ is served by vehicle $k$, and these quantities must be coupled to whether vehicle $k$ actually visits customer $i$. The operator grafts the customers of one active vehicle onto another capacity-tightened vehicle and deactivates the original vehicle's route. The resulting $s_i^-$ is infeasible when visit--delivery coupling is enforced, because the tightened vehicle cannot carry the combined demand. If the regenerated model allows a deactivated vehicle to deliver positive demand without visiting the customer, the same solution can be incorrectly accepted. This exposes models that encode aggregate customer demand satisfaction but omit the coupling between routing variables and split-delivery quantities.

\paragraph{Time window violation} targets hard time window constraints $a_i\le\tau_i^k\le b_i$. It swaps adjacent customers in a feasible route while preserving the customer set, vehicle assignment, and load, and then sets narrow windows around the legal arrival times in the original route. The resulting $s_i^-$ is retained only if the swapped route violates at least one hard time window under the same windows. For PDPTW, the operator is precedence-aware and avoids swaps that directly invert a pickup-delivery pair, so that the violation remains attributable to time windows rather than precedence. Soft-time-window variants do not use this operator because time-window violations are penalized rather than infeasible.

\begin{table*}[t]
\centering
\small
\begin{tabularx}{\textwidth}{lXX}
\toprule
Variant & Structural perturbations & Parameter perturbations \\
\midrule
TSP & Base pair & -- \\
CVRP & Base pair & Capacity overload \\
MDVRP & Base pair & Capacity overload \\
MCVRP & Base pair & Capacity overload; compartment overload \\
PDP & Base pair; precedence violation; split pickup-delivery pair & Capacity overload \\
OVRP & Base pair & Capacity overload \\
DCVRP & Base pair & Capacity overload; distance overrun \\
VRPB & Base pair; backhaul order violation & Capacity overload \\
SDVRP & Split delivery base pair & Quarantine overload; phantom delivery \\
VRPHTW & Base pair & Capacity overload; time window violation \\
VRPSTW & Base pair & Capacity overload \\
HFVRP & Base pair & Capacity overload; heterogeneous fleet overload \\
HFMDVRP & Base pair & Capacity overload \\
HFMCVRP & Base pair & Capacity overload; compartment overload \\
HFVRPHTW & Base pair & Capacity overload; heterogeneous fleet overload; time window violation \\
MDVRPSTW & Base pair & Capacity overload \\
PDPSTW & Base pair; precedence violation; split pickup-delivery pair & Capacity overload \\
PDPTW & Base pair; precedence violation & Time window violation \\
\midrule
\textbf{\textit{OVRPHTW}} & Base pair & Capacity overload; time window violation \\
\textbf{\textit{MCVRPSTW}} & Base pair & Capacity overload; compartment overload \\
\textbf{\textit{DCVRPHTW}} & Base pair & Capacity overload; distance overrun; time window violation \\
\bottomrule
\end{tabularx}
\caption{Perturbation operators enabled for the 21 variants. Base pair denotes the remove-customer and subtour-cycle perturbations. Split-delivery base pair denotes the SDVRP-specific remove-customer and isolated-subtour perturbations. Structural perturbations modify only the visiting sequence, whereas parameter perturbations additionally tighten one target resource family.}
\label{tab:attack_index}
\end{table*}

\subsection{Instance Relaxation and Assembly}
\label{app:relaxation_assembly}
Resource-bound perturbations require an open interval between the resource usage of the feasible solution $s^+$ and that of the corresponding one-constraint-violating solution $s_i^-$. After obtaining $s^+$, we therefore apply one-way relaxation to the bounded resource families present in each variant, such as vehicle capacity, route-distance limits, and time windows. Capacities and route-distance limits are widened relative to the load or route length of $s^+$, while time-window variants receive additional slack around the service times induced by $s^+$. This step only enlarges the feasible region, so $s^+$ remains feasible while later operator-specific tightening has room to place the target bound between the usage of $s^+$ and $s_i^-$.

Let $I^*$ denote the relaxed instance used as the sample instance before perturbation-specific tightening. Structural perturbations are evaluated directly on $I^*$, since their violations are induced by route structure rather than numerical resource bounds. For each resource-bound operator $a$, we create a perturbation-specific copy $I_a^*$ and tighten only the intended resource bound on that copy. Non-target resources may be adjusted only to avoid confounding violations, ensuring that $s_i^-$ targets the intended constraint family. We retain a labeled solution only when $C_{\mathrm{con}}$ verifies $s^+$ as feasible and the corresponding $s_i^-$ as infeasible on the same instance: $I^*$ for structural perturbations and $I_a^*$ for resource-bound perturbations. For each resource-bound sample, its natural-language description and reference concrete code are generated from the same $I_a^*$.

\section{Natural-Language Rewriting and Code Regeneration}
\label{app:synthesis}

This appendix describes Stage 2 of the data synthesis pipeline. We use Gemini-3.1-Pro Preview~\citep{googledeepmind2026gemini} and Claude Opus 4.6~\citep{anthropic2026opus46} as the teacher LLMs in the synthesis pipeline. Within Stage 2, the teacher first rewrites the reference concrete code into natural-language descriptions and then regenerates $C_{\mathrm{con}}^{\mathrm{gen}}$ from each description.

\subsection{Natural-Language Rewriting}
\label{app:three_path_rewriting}

\paragraph{Scenario instantiation.}
The main path rewrites $C_{\mathrm{con}}$ into a natural-language problem statement $p_0$ organized into business background, resources, constraints, and objective. It uses a generate--critique--repair loop: the teacher first produces a draft, a critic then checks numerical completeness, constraint coverage, objective direction, absence of fabrication, ID--attribute binding, and unit plausibility, and a repair call edits only the parts identified as inconsistent. Each call is given a deterministic scene--style hint, described below.

\paragraph{Condensation.}
The first augmentation path removes solver-side or formulation-side wording from $p_0$ while preserving the optimization problem. Typical deletions include self-loop prohibitions, explicit in-degree or out-degree formulations, subtour-elimination wording, and explanatory references to virtual depots or sink nodes when they are only modeling devices. Two prompt variants are used: a permissive version that allows such deletions, and a stricter version that forces virtual-node wording to be replaced by natural route-termination statements. Outputs that fail static fidelity checks are repaired through minimal local edits.

\paragraph{Index rewriting.}
The second augmentation path changes node identifiers in $p_0$ or $p_{\mathrm{cond}}$ while preserving all other semantics. The target style is deterministically selected from three options: consecutive one-based integer identifiers, consecutive integer identifiers starting from 2, or uppercase letter identifiers. The rewrite call must also return a one-to-one \texttt{node\_id\_map}. This map is required downstream because $s^+$ and $\{s_i^-\}$ are constructed on $I^*$ using source node IDs; the verifier must translate the reference solutions into the rewritten naming space before applying $\mathrm{Inj}$.

\paragraph{Scene and style configuration.}
\label{app:scene_config}

Scenario instantiation appends a deterministic scene--style hint to the rewriting prompt. The scene pool contains 60 entries and the style pool contains three entries. Both are selected from a seed derived from the sample triple \((\text{variant}, \text{seed}, \text{profile})\), so rerunning the same sample yields the same scene and style. The template contains no additional random perturbation. The unit hints are drawn from a small set of base unit families, including tons, kilograms, liters, kWh, seat-kilometers, and order-equivalents, sometimes with domain-specific modifiers. The three style instructions are: use a concise academic-problem style similar to a benchmark or methods paper; use a formal optimization-problem style with stable ordering from background to data, constraints, and objective; and use precise conservative wording that avoids storytelling and remains easy to parse into a mathematical model.

\subsection{Code Regeneration}
\label{app:code_regeneration_flow}

Each problem description $p\in\{p_0,p_{\mathrm{cond}},p_{\mathrm{idx}}\}$ is then sent to the code-regeneration prompt to obtain $C_{\mathrm{con}}^{\mathrm{gen}}$. The same code-generation instruction template and output protocol are used when constructing SFT examples, during GRPO rollouts, and in final evaluation.

\section{Constraint-Injection Encoding Details}
\label{app:injection_details}

This appendix provides implementation details for the $\mathrm{Inj}$ check introduced in Section~\ref{sec:rejection}. Given $(C_{\mathrm{con}}^{\mathrm{gen}},s)$, the verifier executes $C_{\mathrm{con}}^{\mathrm{gen}}$ to obtain the Gurobi model, copies the model, replaces the objective with a constant, and appends the solution-encoding constraints $\Phi(s)$. The resulting model is solved only as a feasibility query; its objective value is ignored.

\paragraph{Variable parsing and index alignment.}
The verifier searches for routing variables named \texttt{x}, following the output protocol in Appendix~\ref{app:all_prompts}. It supports both two-dimensional arc variables and three-dimensional arc--vehicle variables, provided that their indices follow the arc-first convention specified by the output protocol. For index-rewritten statements, the recorded \texttt{node\_id\_map} is used to translate the routes in $s$ back to the internal node indices expected by $C_{\mathrm{con}}^{\mathrm{gen}}$. If the routing variables or node indices cannot be matched reliably, the case is not used as a valid $\mathrm{Inj}$ verdict and the sample fails rejection sampling.

\paragraph{Homogeneous-fleet injection.}
For homogeneous fleets, $s^+$ is encoded through a vehicle-agnostic customer--customer projection: customer--customer arcs appearing in $s^+$ are required, customer--customer arcs absent from $s^+$ are forbidden, customer self-loops are blocked, and intentionally unvisited customers have their incoming arcs blocked. Vehicle identities remain free. For each $s_i^-$, the verifier additionally applies the route-level consistency constraints in Eq.~\eqref{eq:phi_route}, so that the customer--customer edges of the same diagnostic route cannot be distributed across interchangeable vehicles.

\paragraph{Heterogeneous-fleet injection.}
For heterogeneous fleets, the verifier uses the vehicle-specific fixing described in Section~\ref{sec:rejection}. This preserves the vehicle assignment carried by the injected solution, which is necessary when capacities, costs, or resource bounds depend on vehicle type. In this case, both $s^+$ and $\{s_i^-\}$ are injected with vehicle identities fixed.

\paragraph{Depot edges.}
The main injection encoding focuses on customer--customer transitions and does not use depot-departure or depot-return patterns as primary evidence for the injected solution. It therefore does not directly test every depot-structure constraint; such errors are detected by the optimal value check only when they change the optimum. For a small number of one-constraint-violating solutions, the verifier adds solution-specific depot-edge restrictions to prevent degenerate repairs, such as serving an intentionally unvisited customer outside the encoded route structure or reassigning an isolated SDVRP customer to another vehicle. These restrictions preserve the intended solution semantics and are not used as a general test of depot-structure constraints.

\paragraph{Split-delivery solutions.}
For SDVRP, the verifier must preserve the service assignment in the injected solution because customer demand may be split across vehicles. The SDVRP-specific one-constraint-violating solutions in Appendix~\ref{app:attack_catalogue}, including quarantine overload and phantom delivery, therefore prevent the violation from being repaired by redistributing the same customer's demand to another vehicle. This keeps the $\mathrm{Inj}$ query aligned with the targeted capacity or visit--delivery coupling constraint.

\paragraph{Verdict handling.}
The feasible solution $s^+$ passes $\mathrm{Inj}$ when the solution-encoded model is feasible. A one-constraint-violating solution $s_i^-$ passes $\mathrm{Inj}$ when the solution-encoded model is infeasible. Cases with model-construction failures, unsupported variable formats, unreliable node matching, or unresolved solver statuses are treated as verification failures during rejection sampling.

\section{Training Details}
\label{app:training_details}
\paragraph{Implementation environment.}
All optimization solves in data synthesis, verification, training-time reward computation, and evaluation are conducted with Gurobi Optimizer under an academic license. The main SFT and GRPO runs share one LoRA adapter configuration with rank \(r=16\), \(\alpha=32\), and dropout 0.1. All model calls use the shared code-regeneration system prompt in Appendix~\ref{app:all_prompts}.

\paragraph{SFT.}
We apply LoRA SFT to Qwen3-8B on 6{,}797 $(p,C_{\mathrm{con}}^{\mathrm{gen}})$ pairs. Training uses four A100-40GB GPUs, per-device batch size 1, gradient accumulation 2, and global effective batch size 8. We train for 3 epochs with sequence length 8192. The optimizer is AdamW with learning rate \(2\times10^{-4}\), cosine scheduling, warmup ratio 0.1, weight decay 0.01, and maximum gradient norm 1.0.

\paragraph{Frontier filtering.}
For each prompt in the SFT pool, we sample $H=6$ offline rollouts from VRPCoder-SFT under the same decoding settings as online GRPO (temperature 0.4, top-$p$ 0.95), score each rollout with the dual verifier under the GRPO reward weights, and retain only prompts whose within-group reward standard deviation is strictly positive. This yields the 716 frontier prompts used in the main GRPO run.

\paragraph{GRPO.}
Starting from the SFT checkpoint, we train on 716 frontier prompts. Policy training uses three A100-40GB GPUs with DeepSpeed ZeRO-2~\citep{rasley2020deepspeed}, while one additional A100-40GB GPU runs a vLLM~\citep{kwon2023vllm} server for rollouts. The per-device batch size is 1, gradient accumulation is 8, and the global effective batch size is 24. We train for 3 epochs. For each prompt, we sample \(G=8\) rollouts with a maximum completion length of 3072, temperature 0.4, top-\(p\) 0.95, and KL coefficient \(\beta=0.02\). The optimizer is AdamW with \(\beta_2=0.95\), learning rate \(2\times10^{-5}\), cosine scheduling, 5 warmup steps, weight decay 0.01, and maximum gradient norm 0.5.

\paragraph{No-injection ablation.}
For the no-injection ablation, we retrain the SFT checkpoint on 7{,}347 samples. Compared with the main 6{,}797-sample SFT set, the additional 550 samples pass the optimal value check but fail constraint injection and are therefore rejected by the main pipeline. All other SFT hyperparameters are unchanged. For GRPO, we remove \(r_{\mathrm{inj}}\) from the reward and keep only build success and the optimal value check, so the maximum reward becomes 0.7 under the original weights. We then generate new rollouts from the 7{,}347-sample no-injection SFT checkpoint and apply the same positive-variance frontier filtering without injection verification, yielding 855 frontier prompts. Decoding, learning rate, scheduling, LoRA configuration, and effective batch size remain aligned with the main GRPO run.

\section{Constraint-Level Error Analysis}
\label{app:constraint_error_analysis}
Across several rounds of synthesis and regeneration, 7{,}347 candidates passed the optimal value check, among which 550 (7.49\%) failed constraint injection. Among the 450 silent-omission cases, the failures were associated with the base structural pair (268), hard time windows (56), backhaul ordering (44), capacity (30), distance limits (24), compartment capacity (21), precedence (5), and split pickup--delivery coupling (2). Here, the base structural pair denotes the remove-customer and subtour-cycle solutions. Among the 100 spurious-over-constraint cases, the failures involved distance limits (40), vehicle capacity (38), hard time windows (21), and compartment capacity (1). To examine whether these synthesis-time patterns persist after training, Table~\ref{tab:constraint_eval_breakdown} reports final performance by active-module count and evaluation source or task.
\begin{table}[t]
\centering
\small
\setlength{\tabcolsep}{4pt}
\begin{tabular}{llrr}
\toprule
Source or slice & Group & $N$ & Pass@1 (\%) \\
\midrule
Subsets 1--2 & 0 active modules & 25 & 100.00 \\
Subsets 1--2 & 1 active module & 20 & 100.00 \\
Subsets 1--2 & 2 active modules & 191 & 95.81 \\
Subsets 1--2 & 3 active modules & 140 & 95.71 \\
\midrule
TSPLIB & TSP & 16 & 100.00 \\
CVRPLIB & CVRP & 15 & 100.00 \\
Solomon & VRPTW & 11 & 81.82 \\
Cordeau-MDVRP & MDVRP & 7 & 85.71 \\
SDVRPLIB & SDVRP & 13 & 92.31 \\
VRPB & VRPB & 12 & 91.67 \\
\midrule
Cleaned SIRL & TSP/VRP & 50 & 100.00 \\
NLCO & TSP & 50 & 92.00 \\
NLCO & CVRP & 50 & 90.00 \\
NLCO & TSPTW$^\dagger$ & 50 & 74.00 \\
NLCO & PDPTW & 50 & 84.00 \\
\bottomrule
\end{tabular}
\caption{VRPCoder-RL Pass@1 by active-module count and evaluation source or task. $\dagger$ denotes a variant absent from training.}
\label{tab:constraint_eval_breakdown}
\end{table}

Hard time windows appear in both synthesis failure modes and remain challenging in final evaluation: VRPCoder-RL obtains 81.82\% Pass@1 on Solomon VRPTW and 74.00\% on the unseen NLCO TSPTW task. Aggregating Subsets 1 and 2 by variant complexity, VRPCoder-RL achieves 95.81\% and 95.71\% Pass@1 on variants with two and three active constraint modules, respectively.

\begin{table*}[t]
\centering
\small
\begin{tabularx}{\textwidth}{rXXXX}
\toprule
ID & Scene & Depot term & Customer term & Vehicle term \\
\midrule
1 & Classical Distribution Planning & distribution center & customer & vehicle \\
2 & Retail Store Replenishment & warehouse & retail store & delivery truck \\
3 & E-commerce Order Fulfillment & fulfillment center & delivery address & delivery van \\
4 & Wholesale Grocery Supply & regional distribution hub & grocery outlet & supply truck \\
5 & Cold-Chain Food Distribution & cold storage facility & supermarket & refrigerated truck \\
6 & Dairy Product Delivery & dairy processing plant & convenience store & insulated vehicle \\
7 & Urban Parcel Last-Mile Delivery & local sorting station & residential address & courier van \\
8 & Meal Delivery Dispatch & cloud kitchen hub & diner location & delivery rider \\
9 & Factory Parts Supply & parts warehouse & assembly workstation & AGV cart \\
10 & Construction Material Delivery & building material yard & construction site & flatbed truck \\
11 & Pharmaceutical Distribution & pharma distribution center & hospital pharmacy & medical transport vehicle \\
12 & Home Healthcare Visiting & community health center & patient residence & nurse dispatch vehicle \\
13 & Municipal Waste Collection & waste treatment plant & collection point & garbage truck \\
14 & School Bus Routing & school campus & student pickup stop & school bus \\
15 & Power Grid Maintenance & maintenance depot & substation & service crew vehicle \\
16 & Telecom Equipment Repair & telecom operations center & base station site & field engineer van \\
17 & Agricultural Produce Collection & produce purchasing station & farm & collection truck \\
18 & Livestock Feed Distribution & feed mill & livestock farm & bulk feed truck \\
19 & Disaster Relief Supply & relief staging area & shelter site & relief vehicle \\
20 & Hazardous Waste Transport & licensed disposal facility & industrial generator site & hazmat transport vehicle \\
21 & Urban Logistics Delivery & distribution center & delivery address & freight truck \\
22 & Food Delivery Dispatch & delivery station & merchant/customer & delivery rider \\
23 & Express Parcel Pickup & courier hub & pickup point & pickup van \\
24 & Furniture Last-Mile Delivery & furniture warehouse & customer apartment & moving van \\
25 & Laundry Pickup and Delivery & laundry processing center & hotel/household & laundry van \\
26 & Fresh Food Cold Chain & cold storage warehouse & grocery supermarket & refrigerated truck \\
27 & Seafood Cross-City Transport & seafood wholesale market & seafood restaurant & live fish transport truck \\
28 & Bakery Product Distribution & central bakery & cafe/pastry shop & bread delivery van \\
29 & Frozen Meal Delivery & frozen food warehouse & vending kiosk & freezer van \\
30 & Agricultural Product Collection & produce purchasing station & farm plantation & farm transport vehicle \\
31 & Flower Wholesale Distribution & flower auction center & florist shop & climate-controlled van \\
32 & Vending Machine Restocking & restocking warehouse & vending machine location & restocking van \\
33 & Bookstore Chain Replenishment & book distribution center & bookstore branch & delivery truck \\
34 & Bike-Sharing Battery Swap & battery swap station & bike parking spot & maintenance dispatch vehicle \\
35 & Ride-Hailing / Custom Bus & vehicle hub & boarding stop & shuttle bus \\
36 & Drone Last-Mile Delivery & drone base station & community receiving tower & logistics drone \\
37 & Autonomous Vehicle Delivery & AGV dispatch station & office building drop-off & AGV delivery robot \\
38 & EV Charging Truck Dispatch & mobile charging depot & stranded EV location & mobile charger truck \\
39 & Medical Supply Delivery & pharmaceutical distribution center & community clinic/hospital & medical cold chain vehicle \\
40 & Blood Bank Distribution & central blood bank & surgical hospital & blood transport vehicle \\
\bottomrule
\end{tabularx}
\caption{Scene hint pool, part I. Entries 1--40 of the 60-entry scene pool.}
\label{tab:scene_examples_part1}
\end{table*}

\begin{table*}[t]
\centering
\small
\begin{tabularx}{\textwidth}{rXXXX}
\toprule
ID & Scene & Depot term & Customer term & Vehicle term \\
\midrule
41 & Mobile Home Nursing & community health center & patient household & visiting nurse \\
42 & Mobile Vaccination Service & disease control center & temporary vaccination site & mobile medical vehicle \\
43 & Medical Waste Collection & medical waste treatment plant & fever clinic & special collection truck \\
44 & Organ Transport Coordination & transplant coordination center & recipient hospital & emergency medical vehicle \\
45 & Municipal Waste Collection & landfill site & residential waste station & garbage truck \\
46 & Snow Removal / De-icing & municipal works depot & major traffic intersection & snowplow \\
47 & School Bus Service & school campus & student pickup point & school bus \\
48 & Mobile Library Tour & city central library & remote community branch & mobile library van \\
49 & Street Light Maintenance & city lighting depot & faulty lamppost location & maintenance bucket truck \\
50 & Armored Cash Transport & vault center & bank branch/ATM & armored transport vehicle \\
51 & Factory JIT Parts Delivery & line-side warehouse & assembly line station & AGV transport robot \\
52 & Port Container Dispatch & container yard & berth/cargo ship & straddle carrier \\
53 & Semiconductor Wafer Transport & cleanroom storage & fabrication tool bay & overhead transport robot \\
54 & High-Voltage Grid Inspection & power maintenance base & transmission tower/substation & drone/inspection vehicle \\
55 & Telecom Base Station Repair & telecom operations center & faulty base station & repair engineer \\
56 & Wind Farm Turbine Maintenance & wind farm service center & wind turbine & maintenance crew vehicle \\
57 & Offshore Oil Platform Supply & onshore supply base & offshore drilling platform & offshore supply vessel \\
58 & Mine Ore Transport & ore processing plant & mining site & heavy-duty mining truck \\
59 & Earthquake Relief Distribution & emergency relief tent/depot & displaced persons shelter & rescue helicopter/off-road vehicle \\
60 & Forest Fire Patrol & forestry fire prevention center & key forest observation point & forest patrol vehicle \\
\bottomrule
\end{tabularx}
\caption{Scene hint pool, part II. Entries 41--60 of the 60-entry scene pool.}
\label{tab:scene_examples_part2}
\end{table*}

\section{Evaluation Benchmarks and Protocol}
\label{app:evaluation_protocol}

VRPBench comprises four subsets with 700 total instances, ranging from in-domain synthesis to externally sourced problems. All problem statements and reference answers are manually verified by operations-research experts.

\paragraph{Subset 1: In-domain synthetic.}
This subset contains 310 instances generated by the three-stage data synthesis pipeline on the 18 training variants with a separate set of random seeds. Samples are retained after the same dual verifier procedure used for training data.

\paragraph{Subset 2: Out-of-domain synthetic.}
This subset contains 66 instances generated by the same pipeline on the three held-out variants OVRPHTW, MCVRPSTW, and DCVRPHTW. The individual constraint modules in these variants appear in training, but the corresponding pairwise combinations are never observed during training, making this benchmark a test of compositional generalization.

\paragraph{Subset 3: Public OR libraries.}
This subset contains 74 instances derived from public OR libraries, including TSPLIB~\citep{reinelt1991tsplib}, CVRPLIB~\citep{uchoa2017new}, \citet{solomon1987algorithms}, Cordeau-MDVRP~\citep{cordeau1997new}, SDVRPLIB~\citep{belenguer2000lower}, and VRPB~\citep{toth1997exact}. It covers TSP, CVRP, VRPTW, MDVRP, SDVRP, and VRPB. Since the original public instances often contain tens to hundreds of customers, we extract 5--12-customer subinstances while preserving real coordinates, demands, time windows, multi-depot structure, split delivery, or backhaul structure when present. OR experts then write the natural-language statements after downsampling, ensuring that all numerical and structural constraints remain consistent.

\paragraph{Subset 4: Externally sourced.}
This subset contains 250 instances. The first 200 come from the routing subset of NLCO~\citep{jiang2026nlco}, covering TSP, CVRP, TSPTW, and PDPTW with 50 small instances per type. We make two lightweight adaptations: removing the JSON output-format and reasoning instructions from the original prompts, and adding the annotated vehicle count to CVRP statements because code-generation tasks require the fleet size to be explicit. The surface format and indexing scheme of NLCO are otherwise preserved. The remaining 50 instances come from the cleaned SIRL release~\citep{chen2025sirl}, which aggregates OptMATH~\citep{lu2025optmath}, IndustryOR~\citep{huang2025orlm}, and MAMO~\citep{huang2024mamo}; we extract all TSP and VRP variants, consisting of 9 OptMATH, 2 IndustryOR, and 39 MAMO ComplexLP cases.

\paragraph{Evaluation.}
All four subsets use instance-level Pass@1 with absolute objective tolerance \(10^{-3}\), a 600-second Gurobi time limit per instance, and greedy decoding. Trained models use the shared code-regeneration system prompt in Figure~\ref{fig:prompt_code_regeneration_system_prompt}; OR-LLM baselines use their original prompts and solver settings.

\section{Prompts Used in Experiments}
\label{app:all_prompts}

This appendix collects all prompt texts used in the paper. Placeholders in braces are filled by the data-generation scheduler at runtime.

\begin{figure*}[p]
\centering
\begin{promptbox}{Scenario Instantiation: System Prompt}
You are an operations research expert. Given Gurobi solver code, produce a formal English business problem description suitable as code-generation training data.
(*@\textbf{Data fidelity:}@*)
1. Preserve ALL numerical data exactly (coordinates, demands, capacities, fleet size, time windows, service times, max distances, etc.). Do not omit, invent, or round any number.
2. State the optimization objective and every constraint in business language. No math symbols, variable names, or code jargon.
3. Index preservation (critical): every vehicle, customer, and depot must carry its explicit numeric ID from the code. Write "Truck 0 (capacity 50.0, fixed cost 70.0)" -- never describe entities by type alone without their ID. Never renumber or reorder.
(*@\textbf{Unit consistency:}@*)
4. Demand/capacity values may be fractional (e.g. 3.4). Use continuous units only (tons, kilograms, liters). Never use bare countable nouns (pallets, boxes, items) -- "3.4 pallets" is nonsensical.
5. If the problem has time-window or scheduling constraints, do NOT also use time-based units for demand/capacity -- this creates a logical contradiction. Use a tangible goods unit instead.
Style:
6. Structure: background -> resources/data -> operational constraints -> objective.
7. Plain prose only. No Markdown, no bullets/tables/dialogue/email tone.
8. If a scene hint is given, adapt terminology naturally while preserving the optimization structure.
\end{promptbox}
\begin{promptbox}{Scenario Instantiation: User Prompt}
Generate a natural-language business problem statement from the code below.
===== Scene and Style Instructions =====
{scene_hint}
===== Gurobi Code (with complete data) =====
{code}
===== Critical Numbers That Must Appear =====
Embed each number naturally in context (e.g. "a capacity of 50.0"), do not list them as a bare table:
{critical_numbers_str}
Output the problem description directly:
\end{promptbox}
\caption{Scenario Instantiation Prompts.}
\label{fig:prompt_scenario_instantiation}
\end{figure*}

\begin{figure*}[p]
\centering
\begin{promptbox}{Self-Critique: Critic System Prompt}
You are a strict reviewer. Compare the Gurobi code with the generated NL description for full consistency.
Check these dimensions:
1. Numerical completeness: all critical values (depot/customer/vehicle counts, capacities, demands, coordinates, time windows, etc.) present?
2. Constraint coverage: every code constraint has a corresponding business description?
3. Objective: optimization direction correctly stated?
4. No fabrication: no numbers or conditions invented beyond the code?
5. Index-attribute binding: every entity carries its numeric ID, and ID-to-attribute mappings (ID -> capacity, cost, type) are faithful to the code? Missing or ambiguous IDs = INCOMPLETE.
6. Unit plausibility: demand/capacity units are continuous (not countable nouns) and do not conflict with time-window dimensions?
Output exactly one of:
- COMPLETE
- INCOMPLETE: <discrepancy_1> | <discrepancy_2> | ...
\end{promptbox}
\begin{promptbox}{Self-Critique: Critic User Prompt}
Please compare the following content and determine whether the natural-language description fully and accurately reflects the optimization problem in the code:
===== Gurobi Code (with complete data) =====
{code}
===== Generated Natural-Language Description =====
{nl_description}
Please provide the evaluation result:
\end{promptbox}
\begin{promptbox}{Repair: System Prompt}
You are an operations research expert.
Revise the natural-language problem description based on the review feedback so that it is fully consistent with the original mathematical problem.
Rules:
1. Only modify the parts identified as problematic.
2. Preserve the correct and fluent parts of the original description.
3. Ensure all critical numbers appear accurately after revision.
4. Do not introduce new errors.
\end{promptbox}
\begin{promptbox}{Repair: User Prompt}
Review feedback: {criticism}
Original code and data: {code}
Current natural-language description: {nl_description}
Please output the revised complete natural-language description (only modify the parts identified as problematic; preserve the correct parts):
\end{promptbox}
\caption{Self-Critique and Repair Prompts. Critic and repair prompts used in the generate--critique--repair loop.}
\label{fig:prompt_self_critique_repair}
\end{figure*}

\newtcblisting{promptboxcompact}[1]{
  listing only,
  enhanced,
  colback=white,
  colframe=black!65,
  boxrule=0.4pt,
  arc=1pt,
  boxsep=3pt,
  left=3pt,
  right=3pt,
  top=1pt,
  bottom=0pt,
  middle=1pt,
  before skip=1pt,
  after skip=1pt,
  title=\textbf{#1},
  fonttitle=\bfseries,
  listing options={
    basicstyle=\normalfont\normalsize,
    aboveskip=0pt,
    belowskip=0pt,
    breaklines=true,
    breakatwhitespace=true,
    breakautoindent=false,
    breakindent=0pt,
    columns=fullflexible,
    keepspaces=true,
    showstringspaces=false,
    escapeinside={(*@}{@*)}
  }
}

\begin{figure*}[p]
\centering

\begin{promptboxcompact}{Condensation: Shared System-Prompt Skeleton}
You condense verbose vehicle-routing problem statements into shorter business-requirement descriptions. The rules are specified as follows:
1. Preserve every numeric value exactly as written. Do not delete, alter, round, merge, or restate numbers in a different form.
2. Preserve every explicit entity ID exactly as written. Do not renumber, reorder, relabel, or collapse depots, customers, vehicles, compartments, pickup-delivery pairs, or any other named entities.
3. Preserve every fact that changes the optimization problem: objective terms, route start/end behavior, depot assignment, fleet size, capacities, coordinates, demands, service times, time windows, maximum distance or duration, fixed or travel costs, heterogeneous vehicle attributes, compartment rules, pickup-delivery logic, backhaul ordering, split-delivery permission, and any other operational restriction.
4. Keep the original business scene and terminology. This is condensation, not stylistic rewriting.
5. <See variant-specific block below>
6. Micro-examples of allowed compression: (1) "each node has exactly one incoming arc and exactly one outgoing arc" can usually be shortened to "each customer/location must be visited exactly once" if route start/end behavior is already clear elsewhere. (2) "no self-loops, no arcs into depot 0, no arcs out of sink 13" can be deleted if these are only solver-hygiene details and the route endpoints are already clear.
7. Merge duplicated statements into one concise statement. Remove tutorial tone, repeated reminders, and repeated restatements of the same requirement.
8. Prefer plain prose. Do not output bullets, markdown, JSON, XML, commentary, or surrounding quotation marks.
9. If the source is already concise, make only small cuts.
Output only the condensed problem statement.
\end{promptboxcompact}

\begin{promptboxcompact}{Condensation: Rule 5 for Pilot 1}
5. You MAY delete solver-side or formulation-side wording when it is not real business content. Typical deletable items include self-loop prohibitions, forbidden-arc wording such as "no arcs into the start depot" or "no arcs out of the sink", exact in-degree / out-degree phrasing, explicit subtour-elimination wording, and explanations of why a virtual depot or sink is introduced.
\end{promptboxcompact}

\begin{promptboxcompact}{Condensation: Rule 5 for Pilot 2}
5. You MUST delete ALL virtual-node wording. Specifically remove: (1) Any mention of "virtual depot", "virtual sink", "virtual return node", "dummy node" and their node IDs when used only as route endpoints. (2) Sentences like "Upon completing their routes, all trucks must proceed to Virtual Distribution center N". (3) Replace with natural route termination: "All vehicles return to the depot after completing deliveries" (closed routes) or "Vehicles do not need to return to the depot" (open routes). You may also delete other solver-side wording (self-loop prohibitions, forbidden-arc phrasing, explicit in-degree / out-degree constraints, subtour-elimination wording).
\end{promptboxcompact}

\begin{promptboxcompact}{Condensation: User Prompt}
Variant: {variant}; Condensation task: Turn the following source description into a shorter business-requirement description. Keep the same scene, terminology, numbers, IDs, objective, and operational constraints. You may compress duplicated explanations and solver-side wording such as self-loop or forbidden-arc phrasing when the business meaning is already clear. Do not introduce a new writing style.
Source description: {original_nl}
Output only the condensed problem statement.
\end{promptboxcompact}

\caption{Condensation prompts. The prompt template used for condensation, consisting of a shared system-prompt skeleton, pilot-specific Rule~5 instructions, and the corresponding user prompt.}
\label{fig:prompt_condensation_prompts}
\end{figure*}

\begin{figure*}[p]
\centering
\begin{promptbox}{Index Rewriting: System Prompt}
Rewrite the VRP problem statement by replacing node/location identifiers only.
Rules:
1. First identify every node/location identifier that appears in the source text: depots, customers, pickup nodes, delivery nodes, virtual depots, sink nodes, endpoint nodes, start nodes, and end nodes.
2. Replace all of those node/location identifiers consistently using the requested target style.
3. Create the node-ID mapping yourself from the source text. Do not rely on any external mapping table.
4. Do not change vehicle IDs or any non-ID number.
5. Preserve all constraints and business meaning.
6. Do not add nodes or virtual-node details that are absent from the source.
7. Report the exact mapping you used in the JSON field node_id_map.
8. Do not output markdown, explanations, or commentary.
Output JSON only:
{
  "rewritten_nl": "...",
  "node_id_map": [
    {"source_id": "0", "target_id": "A", "role": "start_depot"},
    {"source_id": "7", "target_id": "B", "role": "virtual_end_depot"}
  ]
}
\end{promptbox}
\begin{promptbox}{Index Rewriting: User Prompt}
Variant: {variant}
Target style: {target_style}
Source description: {original_nl}
Rewrite the source by replacing every node/location identifier that appears in it. Include physical depots, customer nodes, pickup/delivery nodes, and virtual/sink/end depots if they appear. Decide the mapping yourself from the source text and keep it consistent. Preserve everything else. Return JSON only with rewritten_nl and node_id_map. node_id_map must list every source node/location identifier you replaced and the exact target identifier used in rewritten_nl.
\end{promptbox}
\caption{Index Rewriting Prompts.}
\label{fig:prompt_identifier_rewriting}
\end{figure*}

\begin{figure*}[p]
\centering
\begin{promptbox}{Index Rewriting Check: System Prompt}
You check node-identifier rewrites for VRP problem statements.
Return JSON only with these fields:
{
  "passed": true/false,
  "missing_replacements": [],
  "leaked_source_ids": [],
  "added_absent_nodes": [],
  "mapping_issues": [],
  "issues": []
}
Compare the source statement, rewritten statement, and node_id_map directly. Check only node/location/depot/customer/pickup/delivery/virtual/sink/end identifiers. Do not treat vehicle IDs, coordinates, capacities, demands, costs, times, route counts, vehicle counts, or objective values as node IDs. If a source virtual/sink/end node is present, it must be replaced and included in node_id_map. If a virtual/sink/end node is absent from the source, it must not be added in the rewrite. The rewrite should consistently use the requested target style. node_id_map must describe the actual replacements used in rewritten_nl, must be one-to-one for node identifiers, and must not include vehicle IDs or non-node numeric values.
\end{promptbox}
\begin{promptbox}{Index Rewriting Check: User Prompt}
Target style: {target_style}
Source statement: {original_nl}
Rewritten statement: {rewritten_nl}
Reported node_id_map: {node_id_map}
Check whether every node/location identifier that appears in the source statement has been consistently rewritten in the rewritten statement and correctly recorded in node_id_map. Pay special attention to virtual depots, sink nodes, endpoint nodes, start depots, and end depots. Return JSON only.
\end{promptbox}
\caption{Index Rewriting Check Prompts.}
\label{fig:prompt_identifier_rewrite_check}
\end{figure*}

\begin{figure*}[p]
\centering
\begin{promptbox}{Code Regeneration: System Prompt}
You are an operations research expert. Given a natural-language business problem, generate complete Gurobi Python solver code.
Output ONLY code wrapped in a Python fenced code block. No explanations, no commentary.

Code structure:
1. import gurobipy as gp and from gurobipy import GRB.
2. Define def build_model(): containing all model logic. It must call m.optimize() and return the model object.
3. Set m.setParam('OutputFlag', 0) before optimize.
4. Model ALL constraints described in the problem -- omit none, fabricate none.
5. Use exact numerical values from the problem description.

Index convention:
- Use internal integer node indices 0..N-1 for routing variables, where N is the total number of routing nodes: depots, customers, pickup/delivery nodes, and explicit virtual/end nodes if present.
- If the problem already uses natural zero-based node IDs, use them directly as model indices. In this case, depot is usually node 0, customers are the following nodes, and no node ID mapping is needed.
- If the problem uses any other node ID scheme, such as one-based IDs or simple letter IDs, keep those original IDs only in raw data, then map them to internal indices with node_ids = [...], id2idx = {bid: i for i, bid in enumerate(node_ids)}, idx2id = {i: bid for bid, i in id2idx.items()}, and N = len(node_ids).
- Use input order as node_ids order. Convert every node-indexed attribute through id2idx when a mapping is used: depot IDs, customer IDs, coordinates, demands, service times, time windows, pickup-delivery pairs, virtual/sink depots, and start/end nodes.
- Use node IDs exactly as stated in the problem. Do not invent labels such as Customer A, Node A, Location A, Depot A, Outlet 7, or Hub 0.
- Index fidelity: preserve every ID-to-attribute mapping exactly as described, e.g., Truck 0 capacity 30 means vehicle 0 gets capacity 30. Never reorder or reassign.
- Vehicle IDs are not part of this identifier augmentation requirement; keep the vehicle convention 0..K-1 unless the problem explicitly specifies otherwise.
 
Variable format (critical for verification):
- Route variable MUST be named x with name='x' in addVars, indexed arc-first: x[i,j] or x[i,j,k] (i=origin, j=destination, then vehicle/depot). Never x[k,i,j] or x[(i,j),k].
- x MUST span the full node set, e.g., m.addVars(N, N, K, vtype=GRB.BINARY, name='x'). Do NOT pre-filter arcs with tuplelist; forbid invalid arcs via constraints instead.
- When a mapping is used, set N = len(node_ids) and define x over internal integer node indices. Do not use string/letter business IDs directly as x indices.
\end{promptbox}
\begin{promptbox}{Code Regeneration: User Prompt}
{nl_description}
\end{promptbox}
\caption{Shared code-regeneration prompt used to produce $C_{\mathrm{con}}^{\mathrm{gen}}$ during data synthesis, to generate GRPO rollouts, and to evaluate trained models.}
\label{fig:prompt_code_regeneration_system_prompt}
\end{figure*}

\end{document}